\documentclass[10pt,twocolumn,letterpaper]{article}

\usepackage{iccv}
\usepackage{times}
\usepackage{epsfig}
\usepackage{graphicx}
\usepackage{amsmath}
\usepackage{amssymb}
\usepackage{algorithm}
\usepackage[table]{xcolor}
\usepackage{algorithmic}
\usepackage{setspace}
\usepackage{makecell}
\usepackage{multicol}
\usepackage{multirow}
\usepackage{booktabs}
\usepackage{caption} 
\usepackage{color}
\usepackage{arydshln}
\usepackage{float}
\usepackage{subfig}
\usepackage{enumitem}
\usepackage[accsupp]{axessibility} 
\usepackage[pagebackref=true,breaklinks=true,letterpaper=true,colorlinks,bookmarks=false]{hyperref}
\usepackage{authblk}

\newtheorem{theorem}{Theorem}
\newtheorem{lemma}{Lemma}
\newtheorem{assumption}{Assumption}
\newcommand{\argmin}{\mathop{\mathrm{argmin}}}
\def\etal{\textit{et al}.}
\def\ie{\textit{i.e.}}
\def\eg{\textit{e.g.}}
\definecolor{mygray}{gray}{.9}
\iccvfinalcopy % *** Uncomment this line for the final submission

\def\iccvPaperID{3771} % *** Enter the ICCV Paper ID here

\ificcvfinal\fi

\begin{document}

%%%%%%%%% TITLE
\title{Multimodal Knowledge Expansion}

\author[1,2]{Zihui Xue}
\author[1,3]{Sucheng Ren}
\author[1,4]{Zhengqi Gao}
\author[5,1]{Hang Zhao \thanks{Corresponding to hangzhao@mail.tsinghua.edu.cn}}

\affil[1]{Shanghai Qi Zhi Institute,~~$^2$UT Austin}
\affil[3]{South China University of Technology}
\affil[4]{MIT,~~$^5$Tsinghua University}
% \affil[1]{Shanghai Qi Zhi Institute, $^2$UT Austin}
% \affil[3]{South China University of Technology, $^4$MIT, $^5$Tsinghua University}

% \affil[1]{The University of Texas at Austin}
% \affil[2]{South China University of Technology}
% \affil[3]{Massachusetts Institute of Technology}
% \affil[4]{Tsinghua University}
% \affil[5]{Shanghai Qi Zhi Institute}

% arxiv version
% \author{Zihui Xue$^{1}$~~~~~ Sucheng Ren$^{2}$~~~~~ Zhengqi Gao$^{3}$~~~~~ Hang Zhao$^{4,5}$\thanks{Corresponding to hangzhao@mail.tsinghua.edu.cn.}\\
% $^1$The University of Texas at Austin ~~~~~~~~   $^2$South China University of Technology\\
% $^3$Massachusetts Institute of Technology ~~$^4$Tsinghua University ~~$^5$Shanghai Qi Zhi Institute
% }

%\renewcommand\Authands{ and }

\maketitle
\ificcvfinal\thispagestyle{empty}\fi

%%%%%%%%% ABSTRACT
\begin{abstract}
The popularity of multimodal sensors and the accessibility of the Internet have brought us a massive amount of unlabeled multimodal data. Since existing datasets and well-trained models are primarily unimodal, the modality gap between a unimodal network and unlabeled multimodal data poses an interesting problem: how to transfer a pre-trained unimodal network to perform the same task with extra unlabeled multimodal data? In this work, we propose multimodal knowledge expansion (MKE), a knowledge distillation-based framework to effectively utilize multimodal data without requiring labels. Opposite to traditional knowledge distillation, where the student is designed to be lightweight and inferior to the teacher, we observe that a multimodal student model consistently rectifies pseudo labels and generalizes better than its teacher. Extensive experiments on four tasks and different modalities verify this finding. Furthermore, we connect the mechanism of MKE to semi-supervised learning and offer both empirical and theoretical explanations to understand the 
expansion capability of a multimodal student. \footnote[1]{Code is 
available at: \href{https://github.com/zihuixue/MKE}{https://github.com/zihuixue/MKE}}
\end{abstract}

% for instance, RGB-D images and videos. 
% In MKE, a pretrained unimodal teacher network distills information to a multimodal student on the unlabeled multimodal data. %an interesting phenomenon: although our multimodal student is only provided with pseudo labels given by the teacher, it %We term it as knowledge expansion.

%%%%%%%%% BODY TEXT
\section{Introduction}

%\HZ{Explain how large amount of unlabeled multimodal data is accessible: We show two examples above, (a) hardware upgrade; (b) internet videos.} 
% (a) The robots in the first row indicates that with a hardware upgraded with a new sensor, some new unlabeled multimodal data become accessible. (b) The image and video in the second row indicates lots of unlabelled videos can be easily obtained from the Internet.  } 

% Deep learning has been well explored on unimodal models. With the popularity of multimodal sensors, multimodal data becomes prevalent. Then how to best make use of them?
\begin{figure}[thp]
\centering
\includegraphics[width=0.4\textwidth]{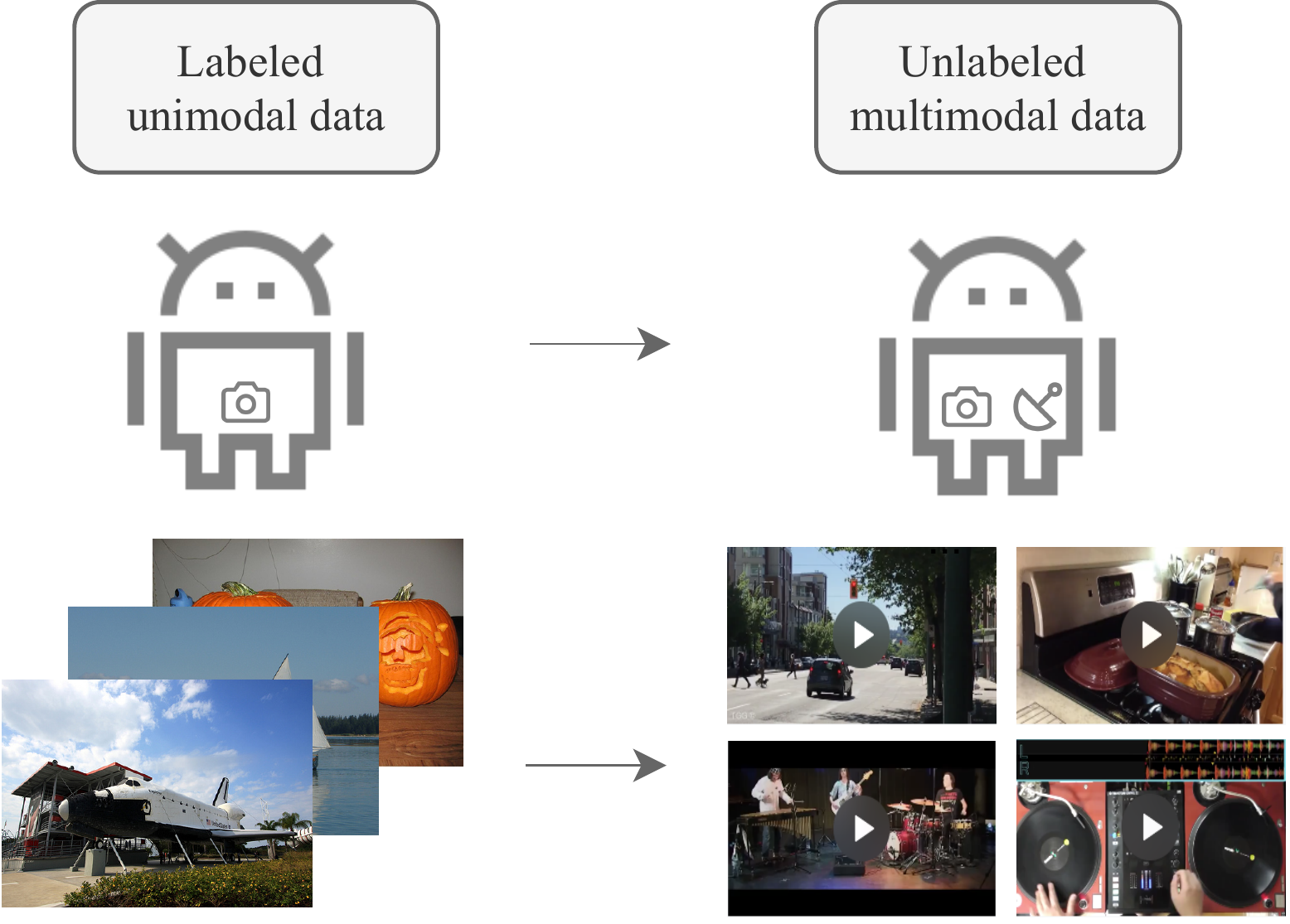} 
\caption{The popularity of multimodal data collection devices and the Internet engenders a large amount of unlabeled multimodal data. We show two examples above: (a) after a hardware upgrade, lots of unannotated multimodal data are collected by the new sensor suite; (b) large-scale unlabeled videos can be easily obtained from the Internet.}
\label{fig:setting}
\end{figure}

Deep neural networks and supervised learning have made outstanding achievements in fields like computer vision~\cite{he2016deep,krizhevsky2012imagenet,simonyan2014very} and computer audition~\cite{hershey2017cnn,yu2016automatic}.
With the popularity of multimodal data collection devices (\eg, RGB-D cameras and video cameras) and the accessibility of the Internet, a large amount of unlabeled multimodal data has become available.
A couple of examples are shown in Figure~\ref{fig:setting}: (a) A unimodal dataset has been previously annotated for the data collected by an old robot; after a hardware upgrade with an additional sensor, the roboticist has access to some new unlabeled multimodal data. (b) Internet videos are abundant and easily accessible. While there are existing unimodal datasets and models for tasks such as image recognition, we further want to perform the same task on unlabeled videos. 
%To utilize these unlabeled multimodal data, 
A natural question arises: \textit{how to transfer a unimodal network to the unlabeled multimodal data?} 
% While they previously annotate unimodal data collected by the old sensor, they now face unannotated multimodal data collected by the new sensor suite as well. 
% \textit{how can we develop models on the unlabeled multimodal data?}

%The modality gap between the unimodal network and the multimodal data poses a major challenge. 
One naive solution is to directly apply the unimodal network for inference using the corresponding modality of unlabeled data. However, it overlooks information described by the other modalities. While learning with multimodal data has the advantage of facilitating information fusion and inducing more robust models compared with only using one modality, developing a multimodal network with supervised learning requires tremendous human labeling efforts.
%labor to annotate the multimodal data.  

% Semi-supervised learning (SSL) has attracted increasing research interest as it mitigates the requirement for labeled data and relaxes human supervision. The goal of SSL is to improve a model's performance by leveraging unlabeled data, in other words, to achieve knowledge expansion from a few labeled data. 

% We introduce a new setting where labeled unimodal data and unlabeled multimodal data are available for training, which differentiates from previous work that focus either on unimodal SSL (i.e., labeled and unlabeled unimodal data) or multimodal SSL (i.e., labeled and unlabeled multimodal data).

% Previous SSL methods made use of unlabeled unimodal data.
% Previous SSL methods limited in using unimodal unlabeled data. A plethora of works [xxx] develop methods to effectively combine both labeled and unlabeled data to boost model performance. Recently, \cite{FixMatch, Noisy student} achieves a remarkable progress on image classification tasks. However, these method faces the confirmation bias problem more or less and their discussion is limited to unlabeled unimodal data only. We introduce a new setting where labeled unimodal data and unlabeled multimodal data are available for training, which differentiates from previous work that focus either on unimodal SSL. Compared with methods that utilize unlabeled unimodal data, the introduction of multiple modalities will bring complementary information to the original modality that can facilitate model training, improve model robustness and alleviate confirmation bias.

\begin{figure}[t]
\centering
\includegraphics[width=0.44\textwidth]{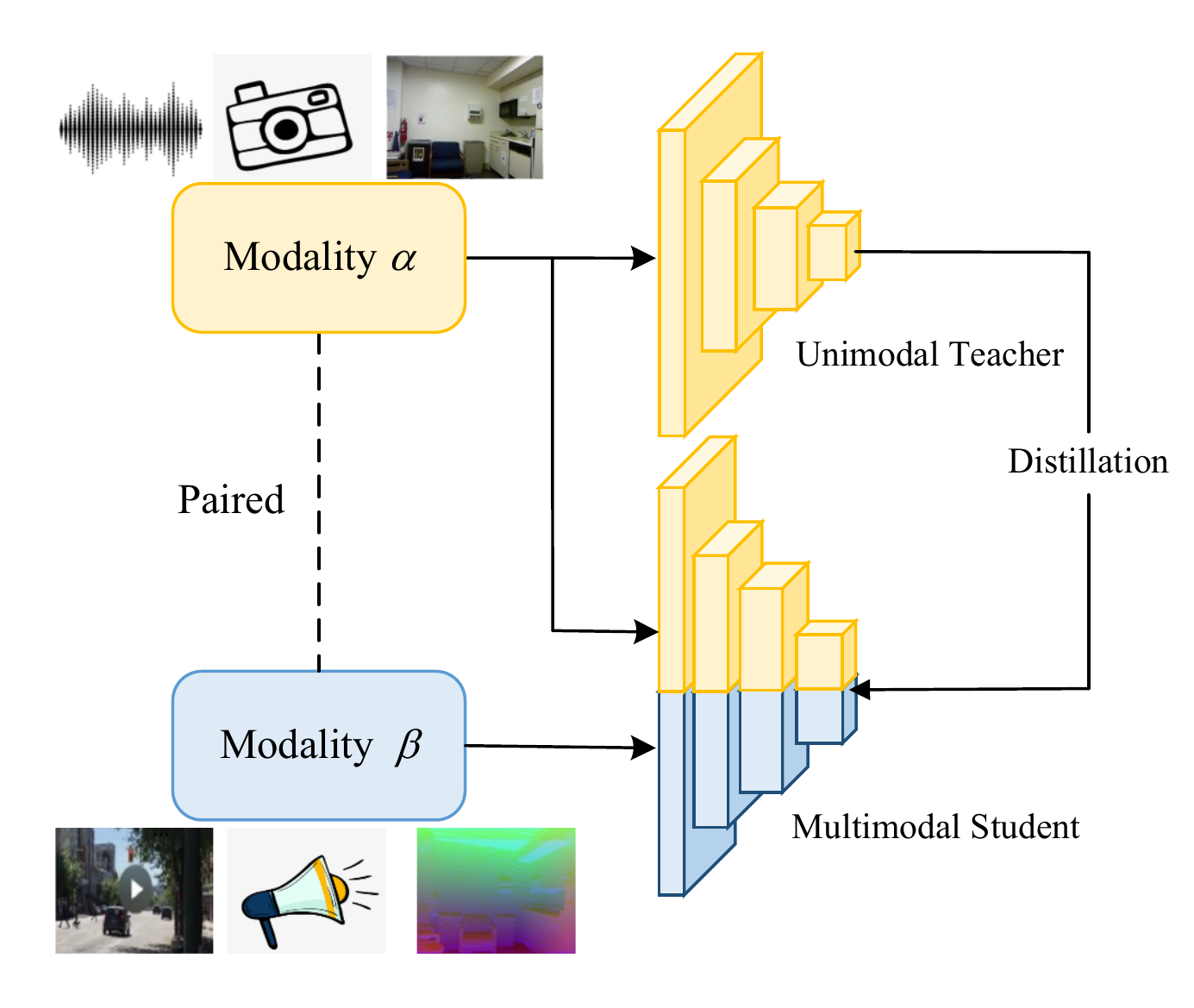} 
\caption{Framework of \textit{MKE}. In \textit{knowledge distillation}, a cumbersome teacher network is considered as the upper bound of a lightweight student network. Contradictory to that, we introduce a unimodal teacher and a multimodal student. The multimodal student achieves \textit{knowledge expansion} from the unimodal teacher.} 
\label{fig:framework} 
\end{figure}

In this work, we propose multimodal knowledge expansion (\textit{MKE}), a knowledge distillation-based framework, to make the best use of unlabeled multimodal data. \textit{MKE} enables a multimodal network to learn on the unlabeled data with minimum human labor (\ie, no annotation of the multimodal data is required). As illustrated in Figure~\ref{fig:framework}, a unimodal network pre-trained on the labeled dataset plays the role of a teacher and distills information to a multimodal network, termed as a student. We observe an interesting phenomenon: our multimodal student, trained only on pseudo labels provided by the unimodal teacher, consistently outperforms the teacher under our training framework. We term this observation as knowledge expansion. Namely, a multimodal student is capable of refining pseudo labels. We conduct experimental results on various tasks and different modalities to verify this observation. We further offer empirical and theoretical explanations to understand the expansion capability of a multimodal student. 

% Opposite to current knowledge distillation methods which typically assume a cumbersome teacher and a lightweight student, where the teacher can be seen as the upper bound of the student/

A closely related setting to ours is semi-supervised learning (SSL), whose goal is to improve a model's performance by leveraging unlabeled data of the same source, including modality. Different from SSL, we aim to develop an additional multimodal network on an unlabeled dataset. Despite the differences in modalities, \textit{MKE} bears some similarity to SSL in terms of the mechanism. We provide a new perspective in addressing confirmation bias, a traditionally bothering problem in SSL. This bias stems from using incorrect predictions on unlabeled data for training and results in marginal performance gain over the original teacher network \cite{arazo2020pseudo}. In SSL, various methods, \ie, data augmentation \cite{fixmatch, uda}, injecting noise \cite{noisystudent}, meta-learning \cite{metapseudolabels} have been proposed to address it. This work provides a novel angle orthogonal to these techniques in alleviating confirmation bias, by resorting to multimodal information. We demonstrate that multimodal inputs serve as a strong regularization, which helps correct inaccurate pseudo labels and overcome the limitation of unimodal networks.

\vspace{10mm}
\section{Related Work}

\subsection{Semi-supervised Learning}
\textbf{Pseudo labeling}, also known as self-training, is a simple and powerful technique in SSL, leading to great improvements on tasks such as image classification \cite{lee2013pseudo, billion,noisystudent,metapseudolabels}, semantic segmentation \cite{zoph2020rethinking, naivestudent} and domain adaptation \cite{zou2019confidence, kumar2020understanding}. One important limitation of pseudo labeling is confirmation bias \cite{arazo2020pseudo}. Since pseudo labels are inaccurate, the student network may potentially learn these mistakes. Various works have been proposed to alleviate this bias \cite{zou2019confidence,arazo2020pseudo,noisystudent,metapseudolabels} while their discussion is limited to unimodality. \textbf{Consistency regularization} is another important brand of SSL. Based on model smoothness assumption, model predictions are constrained to be invariant to small perturbations of either inputs or model hidden states. A series of works have been proposed on producing random perturbations, such as using an exponential moving average of model parameters \cite{tarvainen2017mean}, data augmentation \cite{uda, fixmatch}, dropout \cite{bachman2014learning, noisystudent} or adversarial perturbations \cite{miyato2018virtual}. Recent works \cite{Mixmatch, remixmatch, fixmatch} combine consistency regularization with pseudo labeling together and demonstrate great benefits.

\subsection{Cross-modal Distillation}
Knowledge distillation (KD) \cite{KD} is an effective technique in transferring information from one network to another. KD has been broadly applied to model compression, where a lightweight student network learns from a cumbersome teacher network \cite{kdoverview, KD, yang2019training,romero2014fitnets,tung2019similarity}. Another important application of KD is cross-modal distillation, where a teacher network transfers knowledge from one modality to a student learning from another modality. Various works have been proposed along this direction \cite{cmd, hoffman2016cross, aytar2016soundnet, owens2016ambient, arandjelovic2017look,zhao2018through, cmkd}. Gupta \etal 
~\cite{cmd} proposes a framework that transfers supervision from labeled RGB images to unlabeled depth and optical flow images. SoundNet \cite{aytar2016soundnet} learns sound representations from well established visual recognition models using unlabeled videos. Zhao \etal~\cite{zhao2018through} introduces an approach that estimates human poses using radio signals with cross-modal supervision signals provided by vision models.

\subsection{Multimodal Learning} Models fusing data from multiple modalities has shown superior performance over unimodal models in various applications, for instance, sentiment analysis \cite{sentiment1, sentiment2}, emotion recognition \cite{emotion1, emotion2}, semantic segmentation \cite{feng2020deep,valada2019self, sagate, rgbchannel} and event classification \cite{abavisani2020multimodal}. One recent work \cite{huang2021makes} provides theoretical justifications on the advantages of multimodal learning over learning with single modality.

\renewcommand\arraystretch{1.0}
\begin{table}[!thbp]
\centering
\scalebox{0.75}{
\begin{tabular}{ccccccc}
\toprule

\multirow{3}*{Related works} & \multicolumn{4}{c}{Train} & \multicolumn{2}{c}{Test} \\

~ & \multicolumn{2}{c}{labeled}  & \multicolumn{2}{c}{unlabeled}  & & \\
~ & UM & MM  & UM & MM  & UM & MM \\

\midrule
Semi-supervised learning & $\checkmark$ & & $\checkmark$ & & $\checkmark$ & \\
Cross-modal distillation & $\checkmark$ & & & $\checkmark$ & $\checkmark$ & \\
Supervised multimodal learning & & $\checkmark$ & & & &$\checkmark$ \\
MKE (ours) & $\checkmark$ & & &$\checkmark$  & & $\checkmark$ \\ 

\bottomrule
\end{tabular}}
\caption{Comparison of our data assumption with prior works. UM and MM denotes unimodality and multimodality respectively.}
\label{tab:setting}
\end{table}

We compare our problem setting with prior works in Table \ref{tab:setting}. SSL adopts data from the same modality. Cross-modal distillation has the same training data assumption as us while they focus on testing with unimodal data only. Supervised multimodal learning does not take unlabeled data into consideration. Contrary to them, this work discusses a novel and practical scenario where only labeled unimodal and unlabeled multimodal data are available. 

\section{Approach}
\subsection{Multimodal Knowledge Expansion}

%\textbf{Notations.} While our framework is generalizable to multiple modalities, for conciseness we limit our discussion to two modalities when speaking of multimodality and refer to them as $a$ and $b$, respectively. Consider a classification problem. We use $x_i^a, x_i^b$ to denote inputs from modality $a$ and $b$. Corresponding label $y_i \in \{1, 2, ... , K\}$. 

%\textbf{Problem Formulation.} Assume that we are given a labeled unimodal dataset $D_l=\{(x_i^a,y_i)\}_{i=1}^N$ and an unlabeled dataset $ D_u = \{(x_i^a, x_i^b)\}_{i=1}^M$. Our goal is to train a network $f(x;\theta)$ that could accurately predict the label $y$ when its feature $x$ is given. The network's performance will be tested on a separate set $D_{test} = \{(x_i^a, x_i^b, y_i)\}_i$ 

\noindent\textbf{Problem formulation.} Without loss of generality, we limit our discussion to two modalities, denoted as $\alpha$ and $\beta$, respectively. We assume that a collection of labeled unimodal data $D_l=\{(\mathbf{x}_i^\alpha,\mathbf{y}_i)\}_{i=1}^N$ is given. Each sample input $\mathbf{x}_i^\alpha$ has been assigned a one-hot label vector $\mathbf{y}_i = \{0,1\}^K \in \mathcal{R}^K$, where $K$ is the number of classes. Besides the labeled dataset, an unlabeled multimodal dataset $D_u = \{(\mathbf{x}_i^\alpha, \mathbf{x}_i^\beta)\}_{i=1}^M$ is available. Our goal is to train a network parameterized by $\theta$ (\ie, $\mathbf{f}(\mathbf{x};\theta)$) that could accurately predict the label $\mathbf{y}$ when its feature $\mathbf{x}=(\mathbf{x}^\alpha, \mathbf{x}^\beta)$ is given.

% The `modality gap' between $D_l$ and $D_u$ poses a major challenge when training the network. On one hand, while we are aware of the representation power of multimodal data and intend to adopt multimodal learning, lack of ground truth in $D_u$ seems to make training infeasible. On the other hand, directly utilizing unimodal data $\mathbf{x}^\alpha \in D_l \cup D_u$ for learning is one practical option, but the extra information brought by modality $\beta$ is not incorporated in training and could diminish model performance.

To transfer the knowledge of a labeled unimodal dataset $D_l$ to an unlabeled multimodal dataset $D_u$, we present a simple and efficient model-agnostic framework named multimodal knowledge expansion (\textit{MKE}) in Algorithm \ref{alg:mke}. We first train a unimodal teacher network $\theta_t^\star$ on the labeled dataset $D_l$. Next, the obtained teacher is employed to generate pseudo labels for the multimodal dataset $D_u$, yielding $\tilde{D}_u$. Finally, we train a multimodal student $\theta_s^\star$ based on the pseudo-labeled $\tilde{D}_u$ with the loss term described in Equation (\ref{eq:l_s})-(\ref{eq:l_reg}).

In order to prevent the student from confirming to teacher's predictions (\ie, confirmation bias \cite{arazo2020pseudo}), the loss term in Equation (\ref{eq:l_s})-(\ref{eq:l_reg}) has been carefully designed. It combines the standard pseudo label loss (\ie, Equation (\ref{eq:l_pl})) and a regularization loss (\ie, Equation (\ref{eq:l_reg})). Intuitively speaking, pseudo label loss aims to minimize the difference between a multimodal student and the unimodal teacher, while regularization loss enforces the student to be invariant to small perturbations of input or hidden states. In the context of multimodal learning, the regularization term encourages the multimodal student to learn from the information brought by the extra modality $\beta$, and meanwhile, ensures that the student does not overfit to teacher's predictions based solely on modality $\alpha$. Note that in our implementation, to avoid introducing and tuning one extra hyperparameter $\gamma$ and save computation time, we train the student network with $\theta_s^\star = \argmin_{\theta_s} \frac1 M \sum_{i=1}^M l_{cls}(\tilde{\mathbf{y}}_i, \mathcal{T}(\mathbf{f}_s(\mathbf{x}_i^\alpha, \mathbf{x}_i^\beta;\theta_s))$,  which is equivalent to Equation (\ref{eq:l_s}). The detailed proof is provided in the supplementary material.\\

\begin{figure}[!t]
\centering
\centering
\subfloat[Deeply-colored large dots observed from the X-axis compose $D_l$ and lightly-colored small dots observed from the XY-plane compose $D_u$.]{
\includegraphics[width=0.2\textwidth]{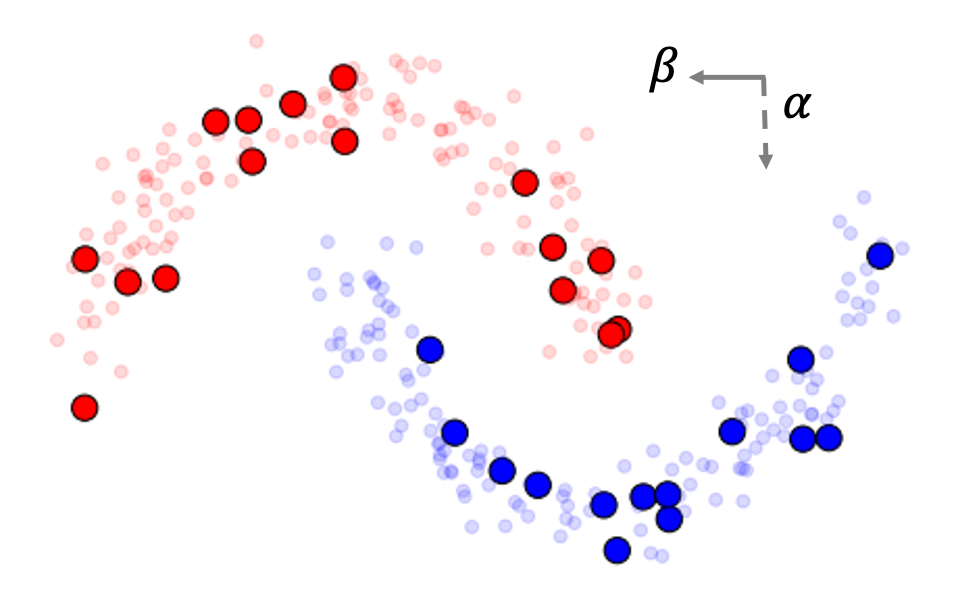}
\label{fig:demo1}
%\caption{fig1}
}
\quad
\subfloat[Train a unimodal classifier on labeled data points. Since only X coordinates of the data in $D_l$ are known, it is natural that the boundary is vertical.]{
\includegraphics[width=0.2\textwidth]{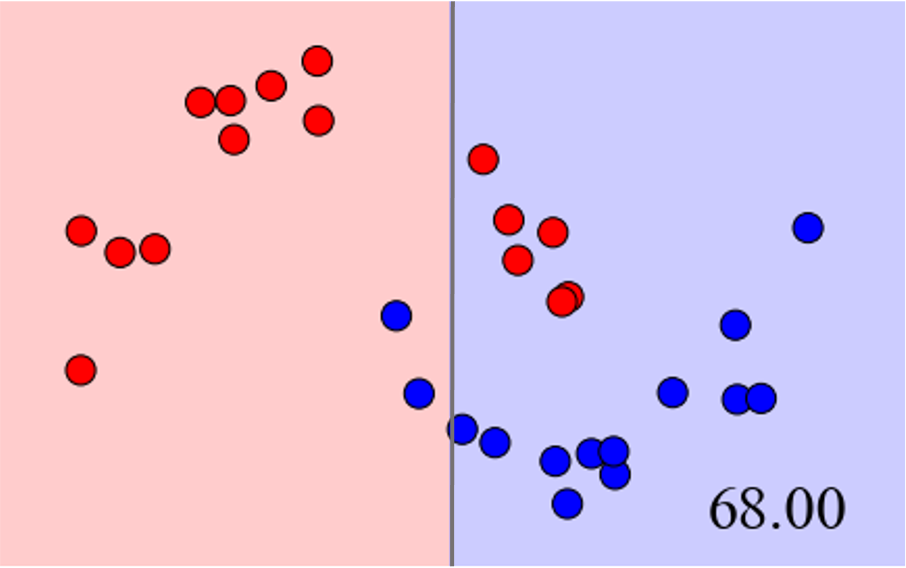}
\label{fig:demo2}
}
\quad
\subfloat[The teacher network generates pseudo labels for unlabeled data]{
\includegraphics[width=0.2\textwidth]{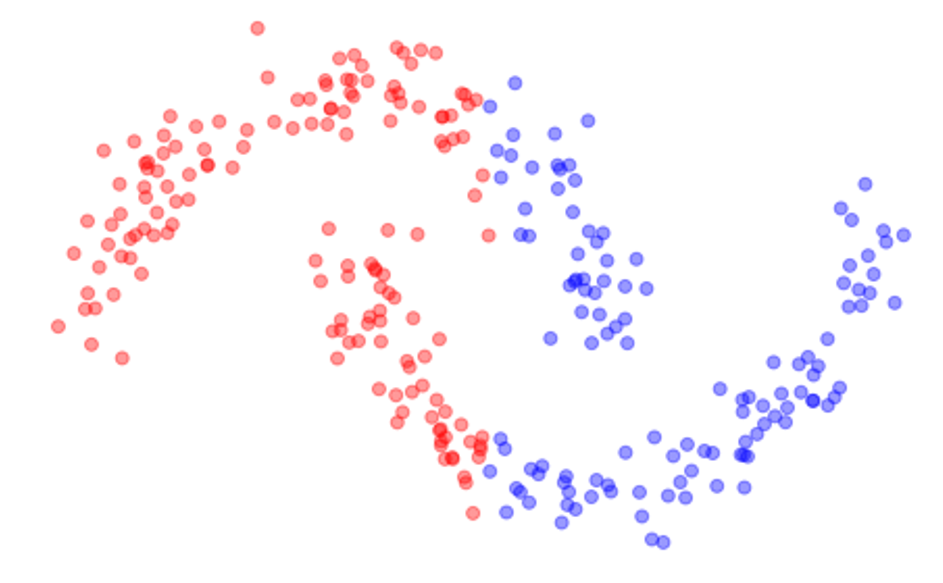}
\label{fig:demo3}
}
\quad
\subfloat[Naive pseudo labeling overfits to inaccurate pseudo labels]{
\includegraphics[width=0.2\textwidth]{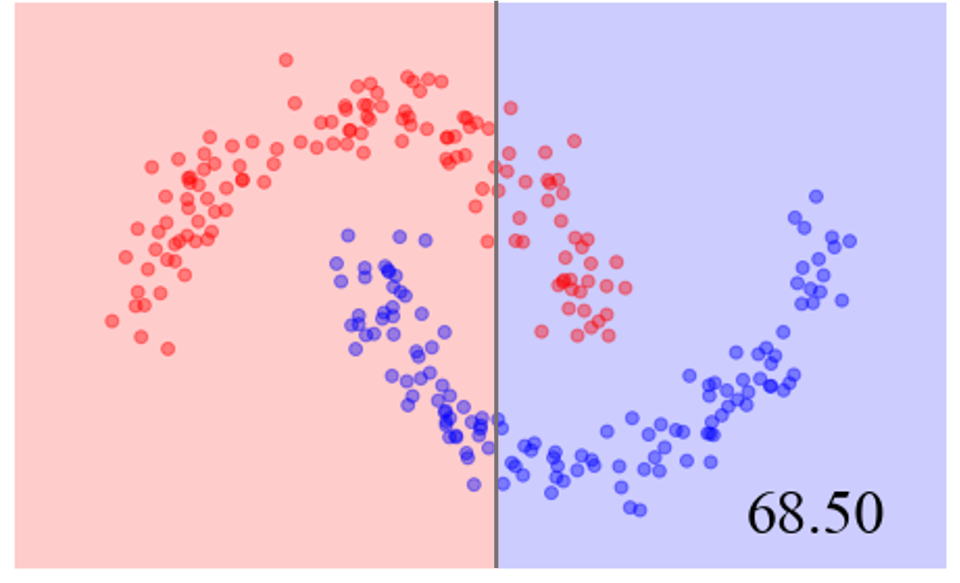}
\label{fig:demo4}
}
\quad
\subfloat[Consistency regularization slightly improves a unimodal student]{
\includegraphics[width=0.2\textwidth]{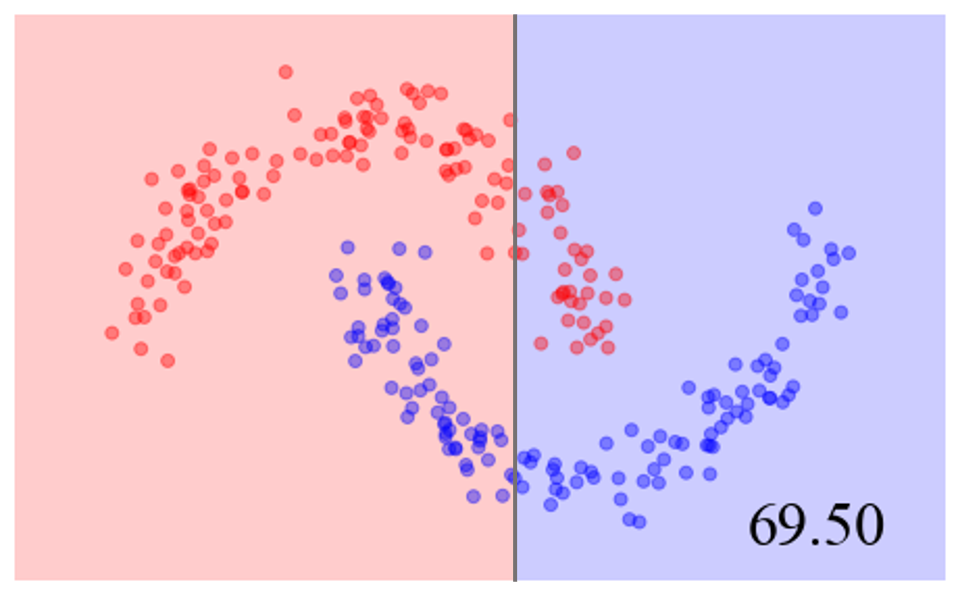}
\label{fig:demo5}
}
\quad
\subfloat[\textit{MKE} greatly improves a multimodal student]{
\includegraphics[width=0.2\textwidth]{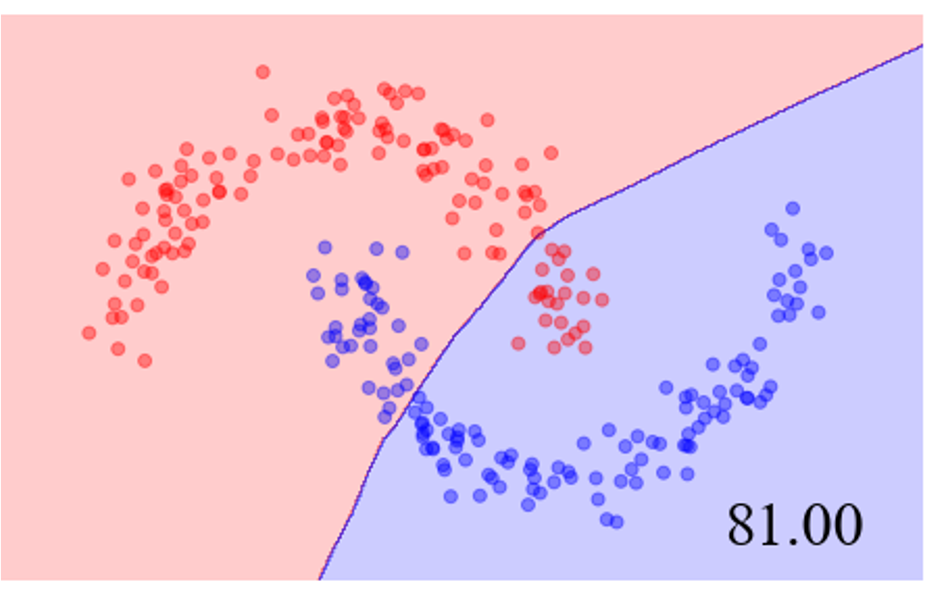}
\label{fig:demo6}
}
\caption{(a)-(c) problem description and illustration of \textit{MKE} using the TwoMoon example; (d)-(f) comparison of naive pseudo labeling, consistency training methods, and the proposed \textit{MKE}. Values in the bottom right corner denotes test accuracy (\%).}
\label{fig:demo}
\end{figure}

\noindent\textbf{An illustrative example.} We consider a variant of the 2D-TwoMoon \cite{arazo2020pseudo} problem shown in Figure \ref{fig:demo1}. The data located at the upper moon and lower moon have true labels 0 and 1, and are colored by red and blue, respectively. The deeply blue- or red-colored large dots compose the labeled unimodal dataset $D_l$, and only their X coordinates are known. On the other hand, $D_u$ consists of all lightly-colored small dots, with both X and Y coordinates available. Namely, modality $\alpha$ and $\beta$ are interpreted as observing from the X-axis and Y-axis, respectively.

\vspace{-2mm}
\begin{algorithm}
\setstretch{1.0}
\caption{multimodal knowledge expansion (\textit{MKE})}
\label{alg:mke}
\begin{algorithmic}
\setlength{\belowdisplayskip}{1pt} \setlength{\belowdisplayshortskip}{1pt}
\setlength{\abovedisplayskip}{2pt} \setlength{\abovedisplayshortskip}{2pt}
\STATE (1) Train a unimodal teacher $\theta_t^\star$ with the labeled dataset $D_l=\{(\mathbf{x}_i^\alpha,\mathbf{y}_i)\}_{i=1}^N$:
\begin{equation}
	\theta_t^\star = \argmin_{\theta_t}\frac1 N \sum_{i=1}^N l_{cls}(\mathbf{y}_i, \mathbf{f}_t(\mathbf{x}_i^\alpha;\theta_t))
\end{equation}

\STATE (2) Generate pseudo labels for $D_u = \{(\mathbf{x}_i^\alpha, \mathbf{x}_i^\beta)\}_{i=1}^M$ by using the teacher model $\theta_t^\star$, yielding the pseudo-labeled dataset $\tilde{D}_u = \{(\mathbf{x}_i^\alpha, \mathbf{x}_i^\beta, \tilde{\mathbf{y}}_i)\}_{i=1}^M$:
\setlength{\belowdisplayskip}{5pt} \setlength{\belowdisplayshortskip}{5pt}
\setlength{\abovedisplayskip}{5pt} \setlength{\abovedisplayshortskip}{5pt}
\begin{equation}
	\tilde{\mathbf{y}}_i = \mathbf{f}_t(\mathbf{x}_i^\alpha;\theta_t^\star), \forall \ (\mathbf{x}_i^\alpha, \mathbf{x}_i^\beta) \in D_u \label{eq:y_tilde}
\end{equation}

\STATE (3) Train a multimodal student $\theta_s^\star$ with $\tilde{D}_u$:
\setlength{\belowdisplayskip}{-1pt} \setlength{\belowdisplayshortskip}{-1pt}
\setlength{\abovedisplayskip}{2pt} \setlength{\abovedisplayshortskip}{2pt}
\begin{equation}
	\theta_s^\star = \argmin_{\theta_s} (\mathcal{L}_{pl} + \gamma \mathcal{L}_{reg})
	\label{eq:l_s}
\end{equation}
\begin{equation}
	\mathcal{L}_{pl} = \frac1 M \sum_{i=1}^M l_{cls}(\tilde{\mathbf{y}}_i, \mathbf{f}_s(\mathbf{x}_i^\alpha, \mathbf{x}_i^\beta;\theta_s) \label{eq:l_pl})
\end{equation}
\begin{equation}
	\begin{aligned}
	\mathcal{L}_{reg} =\sum_{i=1}^M l_{reg} [\mathbf{f}_s(\mathbf{x}_i^\alpha, \mathbf{x}_i^\beta;\theta_s), \mathcal T({\mathbf{f}_s(\mathbf{x}_i^\alpha, \mathbf{x}_i^\beta;\theta_s)})]
	\label{eq:l_reg}
\end{aligned}
\end{equation}
\STATE $l_{cls}$: cross entropy loss for hard $\tilde{\mathbf{y}_i}$ and KL divergence loss for soft $\tilde{\mathbf{y}_i}$.
\STATE $l_{reg}$: a distance metric (\eg, L2 norm).
\STATE $\gamma$: a constant balancing the weight of $\mathcal{L}_{pl}$ and $\mathcal{L}_{reg}$.
\STATE $\mathcal{T}$: a transformation defined on the student model, realized via input or model perturbations (\ie, augmentations, dropout).
\end{algorithmic}
\end{algorithm}
\vspace{-2mm}

We first train a teacher with the labeled unimodal dataset $D_l$. The learned classification boundary is demonstrated in Figure \ref{fig:demo2}. Next, we adopt the learned teacher to generate pseudo labels for $D_u$. As indicated in Figure \ref{fig:demo3}, pseudo labels may be inaccurate and disagree with ground truth: in our toy example, the unimodal teacher only yields 68\% accuracy. As shown in Figure \ref{fig:demo6}, provided with these not-so-accurate pseudo labels, the student could still outperform the teacher by a large margin (\ie, about 13\% more accurate). 
% It implies that the student is capable of denoising inaccurate labels and that knowledge expansion is achieved. 
It presents a key finding in our work: \textit{Despite no access to ground truth, a multimodal student is capable of correcting inaccurate labels and outperforms the teacher network. Knowledge expansion is achieved.}

%Figure \ref{fig:demo6} presents our key observation: Although given pseudo labels of 68\% accuracy, a multimodal student is capable of denoising inaccurate labels and outperforms the teacher by a large margin (i.e., about 80\% - 68\% = 12\%) on test data. It implies that knowledge expansion is achieved by our multimodal student! This somewhat surprising finding further motivates our thinking: \textit{where does the denoising power of a multimodal student come from?}

%This somewhat surprising finding provides a new perspective in semi-supervised learning: besides increasing model capacity and adding consistency regularization, a student network with good representation power can be obtained by resorting to extra information in the input level. Furthermore, the proposed method is orthogonal to current SSL methods. The introduction of a new modality serves as an additional regularization that reduces confirmation bias. 
%
%Going beyond this finding, we investigate reasons behind \textit{MKE} and propose explanations about the `denoising power' of a multimodal student in the following sections. 
% To begin with, we investigate the function of the extra modality $\beta$ from the opposite side. Namely, we assume that the data provided by modality $\beta$ in $D_u$ cannot be observed when training the student. Consequently, our problem reduces to unimodal semi-supervised learning.

\subsection{Rectifying Pseudo Labels}
The somewhat surprising finding about knowledge expansion further motivates our thinking: \textit{where does the expansion capability of a multimodal student come from?} In this section, we will answer this question with the TwoMoon example. 

To start with, we consider directly adopting unimodal SSL for this problem. Namely, given a teacher network $\theta_t^\star$ trained with labeled data $D_l$ and an unlabeled multi-modal dataset $D_u$, the student network takes $\mathbf{x}_i^\alpha \in D_u$ as input. Naive pseudo labeling \cite{lee2013pseudo} uses the following loss to minimize the disagreement between the fixed teacher $\theta_t^\star$ and a student network $\theta_s$:
\begin{equation}
% \begin{aligned}
\mathcal{L}'_{pl}
=\mathbb{E}_{\mathbf{x}_i^\alpha \in D_u}\{ l_{cls}[\mathbf{f}_t(\mathbf{x}_i^\alpha;\theta_t^\star),\mathbf{f}_s(\mathbf{x}_i^\alpha;\theta_s)] \}
% &=& \frac1 M \sum_{i=1}^M l_{cls}[\mathbf{f}_t(\mathbf{x}_i^\alpha;\theta_t^\star),\mathbf{f}_s(\mathbf{x}_i^\alpha;\theta_s)] 
\label{eq:uni-pl}
% \end{aligned}
\end{equation}

However, due to confirmation bias \cite{arazo2020pseudo}, the student network is likely to overfit to incorrect pseudo labels provided by the teacher network, yielding $\mathbf{f}_s(\mathbf{x};\theta_s^\star)$ similar to $\mathbf{f}_t(\mathbf{x};\theta_t^\star)$, if not identical. In the TwoMoon example, we observe that the unimodal student trained with Equation (\ref{eq:uni-pl}) achieves similar performance as its teacher. This is demonstrated in Figure \ref{fig:demo4}. 

To address this bias, we follow the thought of consistency training methods in SSL \cite{miyato2018virtual,uda,fixmatch} and introduce one general regularization loss term to enforce model smoothness:
\begin{equation}
	\mathcal{L}'_{reg} = \mathbb{E}_{\mathbf{x}_i^\alpha \in D_u}\{ l_{reg}[\mathbf{f}_s(\mathbf{x}_i^\alpha;\theta_s), \mathcal{T}'(\mathbf{f}_s(\mathbf{x}_i^\alpha;\theta_s))]\} \label{eq:uni-reg}
\end{equation}
Namely, $\mathcal{L}'_{reg}$ encourages the model to output similar predictions for small perturbations of the input or the model. $\mathcal{T}'(\mathbf{f}_s(\mathbf{x}_i^\alpha;\theta_s))$ denotes transformation applied to unimodal inputs or model hidden states, which can be realized via input augmentation, noise, dropout, etc. As shown in Figure \ref{fig:demo5}, the unimodal student trained with a combined loss of Equation (\ref{eq:uni-pl})-(\ref{eq:uni-reg}) achieves about 69.50\% prediction accuracy. While it indeed outperforms the teacher of 68.00\% accuracy shown in Figure \ref{fig:demo2}, the unimodal student under consistency regularization fails to utilize unlabeled data effectively and only brings marginal improvement. Although confirmation bias is slightly reduced by the regularization term in Equation (\ref{eq:uni-reg}), it still heavily constrains performance of unimodal SSL methods.

% $\mathcal{T}(\mathbf{f}_s(\mathbf{x}_i^\alpha;\theta_s))$ in Algorithm \ref{alg:mke} denotes transformation for the multimodal case

% We use $\mathcal{T}(\mathbf{f}_s(\mathbf{x}_i^\alpha;\theta_s))$ to denote transformation defined on input or hidden states of the model. $\mathcal{T}$ can be realized via various ways, such as data augmentation and dropout.

% Data augmentation \cite{uda, ..} is the most popular choice of input-level transformation in recent years. We take $\mathcal{T}$ to the outermost, making it incorporate model-level transformations as well.
% where $\mathcal{T}$ denotes some sort of transformations. Data augmentation is the most popular choice of input-level transformation, denoted by $\mathbf{f}_s(\mathcal{T}(\mathbf{x}_i^\alpha);\theta_s)$ in \cite{}. In our paper, we take $\mathcal{T}$ to the outermost, making it further incorporate model-level transformations $\mathbf{f}_s(\mathbf{x}_i^\alpha;\mathcal{T}(\theta_s))$, such as feature-level augmentation and dropout \cite{dropout}.
% As shown in Figure \ref{fig:demo5}, the unimodal student trained with this combined loss of Equation (\ref{eq:uni-pl})-(\ref{eq:uni-reg}) achieves about 70.50\% prediction accuracy. It does become better than the teacher shown in Figure \ref{fig:demo2}, but most importantly, still worse than the multimodal student shown in Figure \ref{fig:demo5}. It implies that a unimodal student under consistency regularization is still limited by the representation power of unimodality, and that the extra modality contains information that could enhance model performance. 

Therefore, we turn to multimodality as a solution and resort to the information brought by modality $\beta$. Utilizing both modalities in $D_u$, we substitute unimodal inputs shown in Equation (\ref{eq:uni-pl})-(\ref{eq:uni-reg}) with multimodal ones and derive the loss terms for training a multimodal student:
\begin{equation}
\mathcal{L}_{pl}
= \mathbb{E} \{l_{cls}[\mathbf{f}_t(\mathbf{x}_i^\alpha;\theta_t^\star),\mathbf{f}_s(\mathbf{x}_i^\alpha, \mathbf{x}_i^\beta;\theta_s)]\}
\label{eq:mul-pl}
\end{equation}
\vspace{-6mm}
\begin{equation}
	\mathcal{L}_{reg} = \mathbb{E}\{ l_{reg}[\mathbf{f}_s(\mathbf{x}_i^\alpha, \mathbf{x}_i^\beta;\theta_s), \mathcal{T}(\mathbf{f}_s(\mathbf{x}_i^\alpha, \mathbf{x}_i^\beta;\theta_s))] \}
\label{eq:mul-reg}
\end{equation}
where both expectations are performed with respect to $(\mathbf{x}_i^\alpha, \mathbf{x}_i^\beta) \in D_u$. In fact, Equation (\ref{eq:mul-pl})-(\ref{eq:mul-reg}) reduces to Equation (\ref{eq:l_pl})-(\ref{eq:l_reg}) when $D_u$ is a finite set containing $M$ multimodal samples. As shown in Figure \ref{fig:demo6}, we observe substantial improvement of a multimodal student (\ie, 81.00\% accuracy) over the teacher (\ie, 68.00\% accuracy). It implies that a multimodal student effectively alleviates confirmation bias and leads to superior performance over the teacher.  

%a unimodal student under consistency regularization is still limited by the representation power of unimodality, and that the extra modality further enhances the  multimodal student's performance. 

\begin{figure}[!t]
\centering
\includegraphics[width=0.3\textwidth]{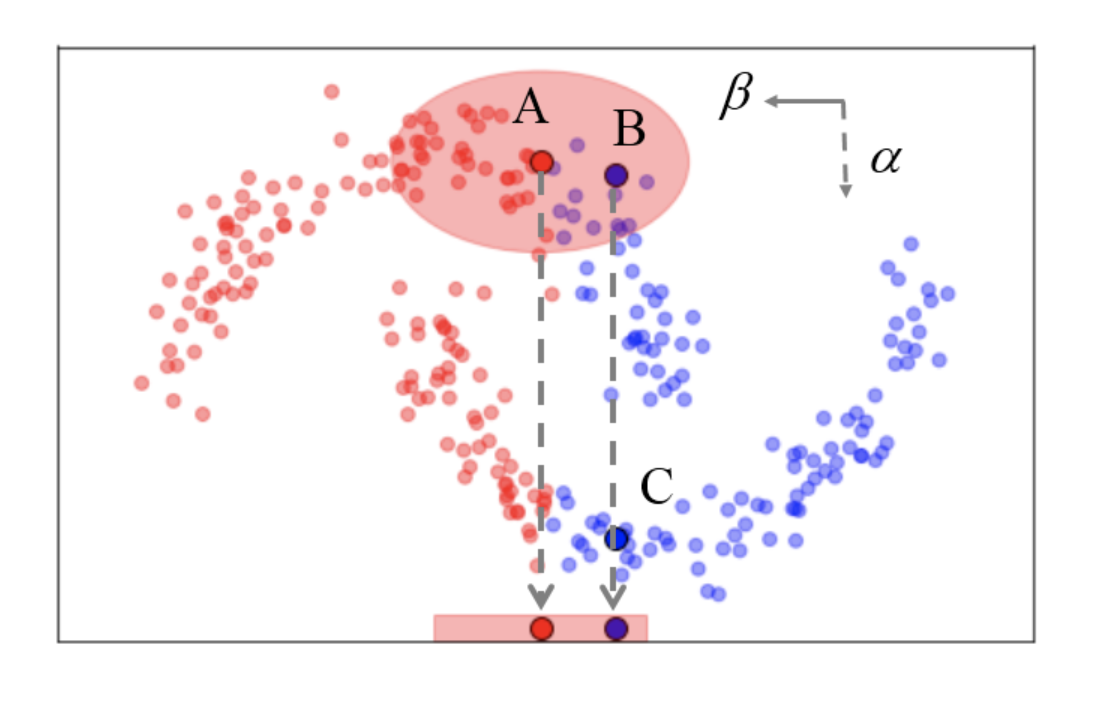} 
\caption{Illustration of the transformation space of one data sample A. The 1-D red line on X-axis corresponds to the transformation space of a unimodal student while the 2-D red circle corresponds to that of a multimodal student.} 
\label{fig:aug} 
\end{figure}

To understand the principles behind this phenomenon, we train one unimodal student with Equation (\ref{eq:uni-pl})-(\ref{eq:uni-reg}) and one multimodal student with Equation (\ref{eq:mul-pl})-(\ref{eq:mul-reg}) on the TwoMoon data. Transformation $\mathcal T$ is defined on model inputs and implemented as additive Gaussian noise. Figure \ref{fig:aug} visualizes the transformation space of one data sample A with both pseudo label and true label being ``red''. Data B is one point that the teacher predicts ``blue'' while its true label is ``red''. The pseudo label and true label of data C are ``blue''.

% To be specific, unimodal transformation $\mathcal T'$ is defined as $\mathcal{T}'(\mathbf{x}_i^\alpha) \triangleq \{{\mathbf{x}_i^\alpha}'| {\mathbf{x}_i^\alpha}' = \mathbf{x}_i^\alpha + \epsilon, \epsilon \sim \mathcal N(0,\sigma^2)\}$, where $\sigma$ is a predefined constant controlling the variance of noise. Similarly, let multimodal transformation $\mathcal T$ be $\mathcal{T}(\mathbf{x}_i^\alpha, \mathbf{x}_i^\beta) \triangleq \{{\mathbf{x}_i^\alpha}', {\mathbf{x}_i^\beta}'| {\mathbf{x}_i^\alpha}' = \mathbf{x}_i^\alpha + \epsilon_1, {\mathbf{x}_i^\beta}' = \mathbf{x}_i^\beta + \epsilon_2, \epsilon_1, \epsilon_2 \sim \mathcal N(0,\sigma^2)\}$. 

When training a unimodal student, we only know the X coordinates of data points, and the transformation space defined by $\mathcal{T}'$ is given by the 1-D red line on X-axis. Under this circumstance, minimizing $\mathcal{L}'_{reg}$ in Equation (\ref{eq:uni-reg}) encourages the unimodal student to predict label ``red'' for the data point located in the red line. This is the case for B, but it will also flip the teacher's prediction for C and make it wrong! The intrinsic reason is that restricted by unimodal inputs, the student network can not distinguish along the Y-axis and mistakenly assumes that C locates near A.

On the contrary, the extra modality $\beta$ helps us see the real distances among A, B, and C. Transformation space of data A in the case of a multimodal student is given by the red circle in Figure \ref{fig:aug}. A multimodal student is guided to predict ``red'' for data falling inside the circle. This time B locates in the transformation space, while C doesn't. Therefore, the multimodal student can correct the wrong pseudo label of data B due to the regularization constraint in Equation (\ref{eq:mul-reg}), and its decision boundary is pushed closer to the ground truth. This example demonstrates that multimodality serves as a strong regularization and enables the student to ``see'' something beyond the scope of its teacher, resulting in knowledge expansion.

\subsection{Theoretical Analysis}
In this section, we provide a theoretical analysis of \textit{MKE}. Building upon unimodal self-training \cite{theory-self-train}, we prove that our multimodal student improves over pseudo labels given by the teacher. 

Consider a $K$-way classification problem, and assume that we have a teacher network pre-trained on a collection of labeled data $D_l$. We further assume a set of unlabeled multimodal data $D_u = \{\mathbf{x}_i = (\mathbf{x}_i^\alpha, \mathbf{x}_i^\beta)\in \mathcal{X}\}_{i=1}^M$ is available, where $\mathcal X = \mathcal X^\alpha \times \mathcal X^\beta$. Let $\mathbf{f}_\star(\mathbf{x};\theta_\star), \mathbf{f}_t(\mathbf{x};\theta_t), \mathbf{f}_s(\mathbf{x};\theta_s)$ denote the ground truth classifier, a teacher classifier, and a student classifier, respectively. Error of an arbitrary classifier $\mathbf{f}(\mathbf{x};\theta)$ is defined as: $Err(\mathbf{f}(\mathbf{x};\theta)) = \mathbb{E}_\mathbf{x}[\mathbf{f}(\mathbf{x};\theta) \neq \mathbf{f}_\star(\mathbf{x};\theta_\star)]$. Let $P$ refer to a distribution of unlabeled samples over input space $\mathcal X$. $P_i$ denotes the class-conditional distribution of $\mathbf{x}$ conditioned on $\mathbf{f}_\star(\mathbf{x};\theta_\star) = i$. We use $\mathcal{M}(\theta_t)\subseteq D_u$ to denote the set of multimodal data that the teacher gives wrong predictions on, \ie, $\mathcal{M}(\theta_t)= \{(\mathbf{x}^{\alpha}, \mathbf{x}^{\beta})|  \mathbf{f}_t(\mathbf{x}^{\alpha};\theta_t) \neq \mathbf{f}_\star(\mathbf{x}^{\alpha};\theta_\star), (\mathbf{x}^{\alpha}, \mathbf{x}^{\beta}) \in D_u \}$. Let $\bar{a} = max_i\{P_i(\mathcal{M}(\theta_t)\}$ refer to the maximum fraction of data misclassified by the teacher network in any class.

% We denote $P_i^\alpha = P(\mathbf{x}|\mathbf{f}_\star(\mathbf{x}^{\alpha};\theta_\star)=i)$ as the class-conditional distribution of $\mathbf{x}^{\alpha}$. $P_i^{\beta}$ follows the same notation, and $P_i = P(\mathbf{x}|\mathbf{f}_\star(\mathbf{x}^{\alpha}, \mathbf{x}^{\beta};\theta_\star)=i)$ refers to distribution of $\mathbf{x}=[\mathbf{x^\alpha}, \mathbf{x^\beta}]$.

%Let $\mathcal{M}(\theta_t)\subseteq D_u$ denote a set of multimodal data that the teacher gives wrong predictions on, i.e., $\mathcal{M}(\theta_t)= \{(\mathbf{x}^{\alpha}, \mathbf{x}^{\beta})|  \mathbf{f}_t(\mathbf{x}^{\alpha};\theta_t) \neq \mathbf{f}_\star(\mathbf{x}^{\alpha};\theta_\star), (\mathbf{x}^{\alpha}, \mathbf{x}^{\beta}) \in D_u \}$. For later convenience, we define $\mathcal{M}^{\alpha}(\theta_t) = \{\mathbf{x}^{\alpha}| \ (\mathbf{x}^{\alpha}, \mathbf{x}^{\beta})\in\mathcal{M}(\theta_t)\}$ and $\mathcal{M}^{\beta}(\theta_t) = \{\mathbf{x}^{\beta}| \ (\mathbf{x}^{\alpha}, \mathbf{x}^{\beta})\in\mathcal{M}(\theta_t)\}$. Let $a_1 = max_i\{P_i(\mathcal{M}^{\alpha}(\theta_t)\}$ refer to the maximum fraction of data misclassified by the teacher in any class from modality $\alpha$\cite{theory-self-train}. Similarly, $a_2\triangleq max_i\{P_i(\mathcal{M}^\beta(\theta_t)\}$. It is obvious that $a_1=a_2$. Let $a = a_1=a_2$.

We first require data distribution $P$ to satisfy the following expansion assumption, which states that data distribution has good continuity in input spaces.
\vspace{-1mm}
\begin{assumption}
$P$ satisfies ($\bar{a}, c_1$) and ($\bar{a}, c_2$) expansion \cite{theory-self-train} on $\mathcal X^\alpha$ and $\mathcal X^\beta$, respectively, with $1<min(c_1,c_2)\leq max(c_1,c_2)\leq\frac{1}{\bar{a}}$ and $c_1c_2>5$.
% $\bar{a}c_1<1, \bar{a}c_2<1, c_1>1, c_2>1, c_1c_2>5$.
\begin{equation}
\label{eq:ac1}
\begin{aligned}
	P_i(&N(V^{\alpha}))  \geq min\{c_1 P_i(V^{\alpha}), 1\}, \\
&\forall \ i \in [K], \forall \ V^{\alpha} \subseteq \mathcal X^{\alpha} \ with \ P_i(V^{\alpha}) \leq \bar{a}
\end{aligned}
\end{equation}
\begin{equation}
\label{eq:ac2}
\begin{aligned}
	P_i(&N(V^{\beta})) \geq min\{c_2 P_i(V^{\beta}), 1\}, \\
&\forall \ i \in [K], \forall \ V^{\beta} \subseteq \mathcal X^{\beta} \ with \ P_i(V^{\beta}) \leq \bar{a}
\end{aligned}
\end{equation}
\end{assumption}
% \vspace{-1mm}
where $N(V)$ denotes the neighborhood of a set $V$, following the same definition as in \cite{theory-self-train}. 

Furthermore, we assume conditional independence of multimodal data in Assumption 2, which is widely adopted in the literature of multimodal learning \cite{ass1-1998, ass2-1998, ass3-2020}. 

% \cite{} states that ground truth labels can be viewed as the generating factor of $\mathcal X^{\alpha}, \mathcal X^{\beta}$. Under this view, a multimodal data pair $(x^{\alpha}, x^{\beta})$ shares the same label: $\mathbf{f}_\star(\mathbf{x^\alpha};\theta_\star) = \mathbf{f}_\star(\mathbf{x^\beta};\theta_\star) = \mathbf{f}_\star(\mathbf{x^\alpha}, \mathbf{x^\beta};\theta_\star)$. 
\vspace{-1mm}
\begin{assumption} 
Conditioning on ground truth labels, $\mathcal X^{\alpha}$ and $\mathcal X^{\beta}$ are independent.
\begin{equation}
\label{eq:cindep}
\begin{aligned}
	P_i(V^{\alpha}&, V^{\beta}) = P_i(V^{\alpha})\cdot P_i(V^{\beta}), \\
& \forall \ i \in [K], \forall \ V^{\alpha} \subseteq \mathcal X^{\alpha}, \forall \ V^{\beta} \subseteq \mathcal X^{\beta}
\end{aligned}
\end{equation}
\end{assumption}
% \vspace{-2mm}

\begin{lemma}
	Data distribution $P$ on $\mathcal X$ satisfies ($\bar{a}, c_1c_2$) expansion.
\end{lemma}
% \vspace{-1mm}
% Multiplying both sides of Equation (\ref{eq:ac1}) and  (\ref{eq:ac2}) and substituting Equation (\ref{eq:cindep}) into the product proves Lemma 1.
Proof of Lemma 1 is provided in the supplementary material. We state below that the error of a multimodal student classifier is upper-bounded by the error of its teacher. We follow the proof in \cite{theory-self-train} to prove Theorem 1.
% \vspace{-1mm}
\begin{theorem}
Suppose Assumption 3.3 of \cite{theory-self-train} holds, a student classifier $\mathbf{f}_s(\mathbf{x}^\alpha, \mathbf{x}^\beta;\theta_s)$ that minimizes loss in Equation (\ref{eq:l_s}) (in the form of Equation 4.1 of \cite{theory-self-train}) satisfies:
\begin{equation}
\begin{aligned}
Err(\mathbf{f}_s(\mathbf{x}^\alpha, \mathbf{x}^\beta;\theta_s)) 
 \leq \frac{4 \cdot Err(\mathbf{f}_t(\mathbf{x}^\alpha;\theta_t))}{c_1c_2-1} + 4\mu 
 \end{aligned}
 \label{eq:mul-bound}
\end{equation}
\end{theorem}
% \vspace{-2mm}
where $\mu$ appears in Assumption 3.3 of \cite{theory-self-train} and is expected to be small or negligible. Theorem 1 helps explain the empirical finding about knowledge expansion. Training a multimodal student $\mathbf{f}(\mathbf{x}^\alpha, \mathbf{x}^\beta;\theta_s)$ on pseudo labels given by a pre-trained teacher network $\mathbf{f}(\mathbf{x}^\alpha;\theta_t)$ refines pseudo labels. 

% \begin{lemma} Under the same training scheme, a multimodal student yields smaller error rate than a unimodal student. 
    % \begin{equation}
% 	Err(\mathbf{f}(\mathbf{x}^\alpha, \mathbf{x}^\beta;\theta_s) \leq Err(\mathbf{f}(\mathbf{x}^\alpha;\theta_s))
% 	\label{eq:umvsmm}
% \end{equation}
% \end{lemma}

In addition, the error bound of a unimodal student $\mathbf{f}_s(\mathbf{x}^\alpha;\theta_s)$ that only takes inputs from modality $\alpha$ and pseudo labels is given by:
\begin{equation}
Err(\mathbf{f}_s(\mathbf{x}^\alpha;\theta_s))
 \leq \frac{4 \cdot Err(\mathbf{f}_t(\mathbf{x}^\alpha;\theta_t))}{c_1-1} + 4\mu 
 \label{eq:uni-bound}
\end{equation}
By comparing Equation (\ref{eq:mul-bound}) and (\ref{eq:uni-bound}), we observe that the role of multimodality is to increase the expansion factor from $c_1$ to $c_1c_2$ and to improve the accuracy bound. This observation further confirms our empirical finding and unveils the role of \textit{MKE} in refining pseudo labels from a theoretical perspective. 
\section{Experimental Results} \label{sec:exp}

To verify the efficiency and generalizability of the proposed method, we perform a thorough test of \textit{MKE} on various tasks: (i) binary classification on the synthetic TwoMoon dataset, (ii) emotion recognition on RAVDESS \cite{ravdess} dataset, (iii) semantic segmentation on NYU Depth V2 \cite{nyudepth} dataset, and (iv) event classification on AudioSet \cite{audioset} and VGGsound \cite{vggsound} dataset. We emphasize that the above four tasks cover a broad combination of modalities. For instance, modalities $\alpha$ and $\beta$ represent images and audios in (ii), where images are considered as a ``weak'' modality in classifying emotions than images. In (iii), modality $\alpha$ and $\beta$ refer to RGB and depth images, respectively, where RGB images play a central role in semantic segmentation and depth images provide useful cues.
~\\

\noindent\textbf{Baselines.}
Our multimodal student (termed as MM student) trained with \textit{MKE} is compared with the following baselines:
\begin{itemize}[topsep=2px,itemsep=-3px]
\item UM teacher: a unimodal teacher network trained on $(\mathbf{x}^\alpha, \mathbf{y}_i) \in D_l$. 
\item UM student: a unimodal student network trained on $(\mathbf{x}^\alpha, \tilde{\mathbf{y}}_i) \in \tilde{D}_u$ (\ie, uni-modal inputs and pseudo labels given by the UM teacher).
\item NOISY student \cite{noisystudent}: a unimodal student network trained on $(\mathbf{x}^\alpha, \mathbf{y}_i) \in D_l \cup (\mathbf{x}^\alpha, \tilde{\mathbf{y}}_i) \in \tilde{D}_u$ with noise injected during training. 
\item MM student (no reg): a multimodal student network trained with no regularization (\ie, Equation (\ref{eq:l_reg}) is not applied during training). % By detaching regularization techniques, we aim to show that our proposed \textit{MKE} and mainstream regularization methods for SSL are orthogonal to each other. Incorporating multimodality won't make the performance gain brought by regularization disappear.
\item MM student (sup): a multimodal student trained on $D_u$ with true labels provided. This supervised version can be regarded as the upper bound of our multimodal student.
\end{itemize}

Since iterative training \cite{noisystudent} can be applied to other baselines and our MM student as well, the number of iterations of a NOISY student is set as one to ensure a fair comparison. We employ different regularization techniques as $\mathcal T$ in Equation (\ref{eq:l_reg}) for the four tasks to demonstrate the generalizability of our proposed methods. Regularization is applied to all baselines identically except for MM student (no reg). 

Furthermore, we present an ablation study of various components of \textit{MKE}, \ie, unlabeled data size, teacher model, hard \textit{vs.} soft labels, along with dataset and implementation details in the supplementary material.

%\begin{itemize}
%	\item Binary Classification: we generate the TwoMoon dataset \cite{twomoon} and interpret two dimensions of data as modality $\alpha$ and $\beta$ respectively.
%	\item Emotion Recognition: we use the Ryerson Audio-Visual Database of Emotional Speech and Song (RAVDESS) \cite{ravdess} dataset. Images and audios correspond to modality $\alpha$ and $\beta$.
%	\item Semantic Segmentation: we conduct experiments on NYU Depth v2 \cite{nyud}. Modality $\alpha$  and $\beta$ refers to RGB images and depth images. 
%	\item Audio and Video Event Classification: we take the mini common set of AudioSet and VGGSound dataset. Audio and video represent modality $\alpha$ and modality $\beta$ respectively.

%\end{itemize}
\subsection{TwoMoon Experiment}
We first provide results on synthetic TwoMoon data. We generate 500 samples making two interleaving half circles, each circle corresponding to one class. The dataset is randomly split as 30 labeled samples, 270 unlabeled samples and 200 test samples. X and Y coordinates of data are interpreted as modality $\alpha$ and $\beta$, respectively.
~\\

% The 2-D nature of data is a natural fit to illustrate our multimodality setting and visualize results. 
% \textbf{Dataset.} We generate 500 samples making two interleaving half circles and the dataset is randomly split as 30 labeled samples, 270 unlabeled samples and 200 test data (i.e., $|D_l|=30$, $|D_u|=270$, and  $|D_{test}|=200$). Note that only one feature (i.e, X coordinates) of labeled data $D_l$ is accessible, while unlabeled data $D_u$ have both X and Y coordinates as features. In other words, X and Y coordinates of data are interpreted as modality $\alpha$ and $\beta$, respectively. 

\noindent\textbf{Baselines \& Implementation.} 
% Besides comparing our multimodal student with the unimodal teacher, we train a unimodal student that could only takes inputs from modality $\alpha$ of unlabeled data $D_u$. Furthermore, we train a multimodal student with unlabeled data $D_u$ by assuming that all true labels are provided. This supervised version of multimodal student can be regarded as the upper bound of our multimodal student.
We implement both the UM teacher and the UM student networks as 3-layer MLPs with 32 hidden units, while the MM student has 16 hidden units. We design three kinds of transformations $\mathcal T = \{\mathcal T_1, \mathcal T_2, \mathcal T_3\}$ used in Equation (\ref{eq:l_reg}): (i) $\mathcal T_1$: adding zero-mean Gaussian noise to the input with variance $v_0$, (ii) $\mathcal T_2$: adding zero-mean Gaussian noise to outputs of the first hidden layer with variance $v_1$, and (iii) $\mathcal T_3$: adding a dropout layer with dropout rate equal to $r_0$. By adjusting the values of $v_0$, $v_1$ and $r_0$, we could test all methods under no / weak / strong regularization. Specifically, higher values indicate stronger regularization.  

% We implement both the UM teacher and the UM student networks as 3-layer MLPs with 32 hidden units, while the MM student has 16 hidden units. NOISY student is not implemented in this small example. By reducing the parameters of a MM student network, we aim to show that its performance gain does not relate to model capacity when compared with a UM student. 

\renewcommand\arraystretch{1.0}
\begin{table}[!htbp]
\centering
\begin{tabular}{cccc}
\toprule
% \midrule
Methods & \multicolumn{3}{c}{Test Accuracy (\%)}\\
\midrule
UM teacher &  \multicolumn{3}{c}{68.00} \\

\midrule
% & \multicolumn{3}{c}{(i) input-level Gaussian noise} \\
 $\mathcal T_1$ & $v_0=0$ & $v_0=1$ & $v_0=2$ \\
\makecell[c]{UM student} & 68.00 & 69.90 & 72.80\\
\makecell[c]{MM student (ours)} & 68.85 & 80.75 & \textbf{83.15}\\
\hdashline
\makecell[c]{MM student (sup)} & 88.05 & 87.35 & 86.95\\
\midrule

%  & \multicolumn{3}{c}{(ii) feature-level Gaussian noise} \\
$\mathcal T_2$ & $v_1=0$ & $v_1=5$ & $v_1=10$ \\
\makecell[c]{UM student} & 68.00 & 68.95 & 70.05 \\
\makecell[c]{MM student (ours)} & 68.85 & 80.00 & \textbf{82.10} \\
\hdashline
\makecell[c]{MM student (sup)} & 88.05 & 87.40 & 86.40  \\
\midrule

%  & \multicolumn{3}{c}{(iii) dropout} \\
$\mathcal T_3$ & $r_0=0$ & $r_0=0.4$ & $r_0=0.8$ \\
\makecell[c]{UM student} & 68.00 & 68.40 & 68.95 \\
\makecell[c]{MM student (ours)} & 68.85 & 73.65 & \textbf{79.20} \\
\hdashline
\makecell[c]{MM student (sup)} & 88.05 & 87.35 & 86.90  \\

\bottomrule
\end{tabular}

% \caption{Results of the TwoMoon experiment. UM and MM denote uni- and multi- modality, respectively. 'sup' denotes a multimodal student trained with true labels in a supervised way.}
\caption{Results of TwoMoon experiment. A MM student significantly outperforms a UM student and teacher under consistency regularization.}
\label{tab:twomoon}
\end{table}

\vspace{3mm}
\noindent\textbf{Results.} Table \ref{tab:twomoon} demonstrates that a MM student under consistency regularization outperforms its unimodal counterpart in all cases of $\mathcal T$. Specifically, a MM student under strong regularization achieves closes results with MM student (sup), as shown in the last column. The small gap between a MM student (trained on pseudo labels) and its upper bound (trained on true labels) indicates the great expansion capability of \textit{MKE}. In addition, we observe better performance of both UM and MM student with increasing regularization strength, demonstrating that consistency regularization is essential in alleviating confirmation bias.

\subsection{Emotion Recognition}
We evaluate \textit{MKE} on RAVDESS \cite{ravdess} dataset for emotion recognition. The dataset is randomly split as 2:8 for $D_l$ and $D_u$ and 8:1:1 as train / validation / test for $D_u$. Images and audios are considered as modality $\alpha$ and $\beta$, respectively. 
~\\

% After validating our idea on synthetic data, we further conduct experiments on real datasets. 

% \textbf{Dataset.} The Ryerson Audio-Visual Database of Emotional Speech and Song (RAVDESS) \cite{ravdess} contains videos and audios of 24 professional actors (12 female, 12 male), vocalizing two lexically-matched statements. It contains 1440 emotional utterances with 8 different emotion classes: neutral, calm, happy, sad, angry, fearful, disgust and surprise. The dataset is randomly split as 2:8 for $D_l$ and $D_u$ and 8:1:1 as train / validation / test for $D_u$. To construct the labeled unimodal dataset $D_l$, we select images every 0.5 second of a video as modality $\alpha$ and train a facial emotion recognition (FER) network as the UM teacher, which classifies emotions based on images. For the unlabeled multimodal dataset $D_u$, we sample images and audios from video clips. We sample images as inputs from modality $\alpha$ in the same way, adopt "Kaiser best" sampling for audios and take Mel-frequency cepstral coefficients (MFCCs) as inputs from modality $\beta$.
\noindent\textbf{Baselines \& Implementation.} For the MM student, we adopt two 3-layer CNNs to extract image and audio features, respectively. The two visual and audio features are concatenated into a vector and then passed through a 3-layer MLP. The UM teacher, UM student and NOISY student are identical to the image branch of a MM student network, also followed by a 3-layer MLP. $\mathcal T$ in Equation (\ref{eq:l_reg}) is implemented as one dropout layer of rate 0.5.
~\\

\noindent\textbf{Results.} As shown in Table \ref{tab:emo}, with the assistance of labeled data and consistency regularization, NOISY student generalizes better than the UM teacher and UM student, achieving 83.09\% accuracy over 80.33\% and 77.79\%. Still, the improvement is trivial. In contrast, our MM student network improves substantially over the original teacher network despite no access to ground truth and leads to 91.38\% test accuracy. The great performance gain can be attributed to additional information brought by audio modality. It demonstrates that \textit{MKE} can be plugged into existing SSL methods like NOISY student for boosting performance when multimodal data are available. Furthermore, regularization helps our MM student yield better performance than the MM student (no reg). More results are presented in the supplementary material.

% by comparing MM student (no reg) with our MM student, we show that \textit{MKE} is orthogonal to mainstream SSL methods that enforce consistency regularization.
% \textbf{Baselines.} Apart from the unimodal teacher and a unimodal student trained with pseudo labels only, we add another baseline of a unimodal student that can access labeled data, termed as noisy student \cite{noisy-student}. Following \cite{noisy-student}, the noisy student is trained with combined loss on both labeled and unlabeled data. Regularization is applied to all unimodal and multimodal architectures, and we add one experiment where we remove the regularization term when training a multimodal student to illustrate the effect of combining multimodality and regularization.

\renewcommand\arraystretch{1.0}
\begin{table}[!htbp]
\centering
\begin{tabular}{cccccc}
\toprule

\multirow{2}*{Methods} & \multicolumn{3}{c}{Train data} & \multicolumn{2}{c}{Accuracy (\%)} \\

~ & \textit{mod} & $D_l$ & $\tilde{D}_u$ & val & test \\

\midrule
UM teacher & $i$ & $\checkmark$ & & 79.67 & 80.33 \\
UM student & $i$ &  &$\checkmark$  & 79.01 & 77.79 \\
NOISY student \cite{noisystudent} & $i$ & $\checkmark$ & $\checkmark$ & 82.54 & 83.09 \\
\makecell[c]{MM student (no reg)} & $i, a$ & & $\checkmark$ & 88.73 & 89.28 \\
\makecell[c]{MM student (ours)} & $i, a$ & & $\checkmark$ & 90.61 & \textbf{91.38} \\
\hdashline
\makecell[c]{MM student (sup)} & $i, a$ &  & $\star$ & 97.46 & 97.35\\
\bottomrule
\end{tabular}
\caption{Results of emotion recognition on RAVDESS.\ \textit{mod}, $i$ and $a$ denote modality, images and audios, respectively. Data used for training each method is listed. $\star$ means that the MM student (sup) is trained on true labels instead of pseudo labels in  $\tilde{D}_u$.}
\label{tab:emo}
\end{table}

%To better quantify the regularization effect that audio modality brings, we report accuracy compared with pseudo labels of each method in the first column. Results indicate that a unimodal student still overfits to pseudo labels while a multimodal student effectively denoises pseudo labels and behaves similarly to MM student (sup) trained with ground truth.

\subsection{Semantic Segmentation}
% In the previous experiment, audio can be understood as a strong modality over RGB images in classifying emotions. In the following sections, we present experimental results where modality $\beta$ is weak compared with modality $\alpha$ and show that our \textit{MKE} suits both cases well. RGB-D semantic segmentation is one example where RGB images (modality $\alpha$ in our setting) play a central role in prediction and depth images (modality $\beta$) provide useful cues. 
We evaluate our method on NYU Depth V2 \cite{nyudepth}. It contains 1449 RGB-D images with 40-class labels, where 795 RGB images are adopted as $D_l$ for training the UM teacher and the rest 654 RGB-D images are for testing. Besides labeled data, NYU Depth V2 also provides unannotated video sequences, where we randomly extract 1.5K frames of RGB-D images as $D_u$ for training the student. Modality $\alpha$ and $\beta$ represents RGB images and depth images.
% \textbf{Dataset.} NYU Depth v2 \cite{nyud} contains 1449 RGB-D images with 40-class labels, where 795 RGB images are used for training a teacher network and the rest 654 images are for testing. Besides labeled data, NYU Depth v2 also provides unannotated video sequences, where we randomly extract 1488 frames of RGB-D images as unlabeled data for training the student.
\renewcommand\arraystretch{1.0}
\begin{table}[!htbp]

\centering
\begin{tabular}{ccccc}
\toprule
\multirow{2}*{Method} & \multicolumn{3}{c}{Train data} & \multirow{2}*{\makecell[c]{Test mIoU \\(\%)}} \\
~ & \textit{mod} & $D_l$ & $\tilde{D}_u$ &  ~ \\
\midrule
UM teacher & $rgb$ & $\checkmark$ & & 44.15  \\
Naive student \cite{naivestudent} & $rgb$ &  &$\checkmark$  & 46.13  \\
NOISY student \cite{noisystudent} & $rgb$ & $\checkmark$ & $\checkmark$ & 47.68  \\
Gupta \etal~ \cite{cmd} & $rgb,d$ & & $\checkmark$ & 45.65  \\
CMKD~ \cite{cmkd} & $rgb,d$ & & $\checkmark$ & 45.25  \\
\makecell[c]{MM student (no reg)} & $rgb, d$ & & $\checkmark$ & 46.14   \\
\makecell[c]{MM student (ours)}& $rgb, d$ & & $\checkmark$ & \textbf{48.88}  \\

\bottomrule
\end{tabular}
\caption{Results of semantic segmentation on NYU Depth V2.\ $rgb$ and $d$ denote RGB images and depth images.}
\label{tab:seg}
\end{table}

\begin{figure*}[!thp]
\centering
\subfloat[RGB images]{
\begin{minipage}[b]{0.15\textwidth}
\centering
\includegraphics[scale=0.2]{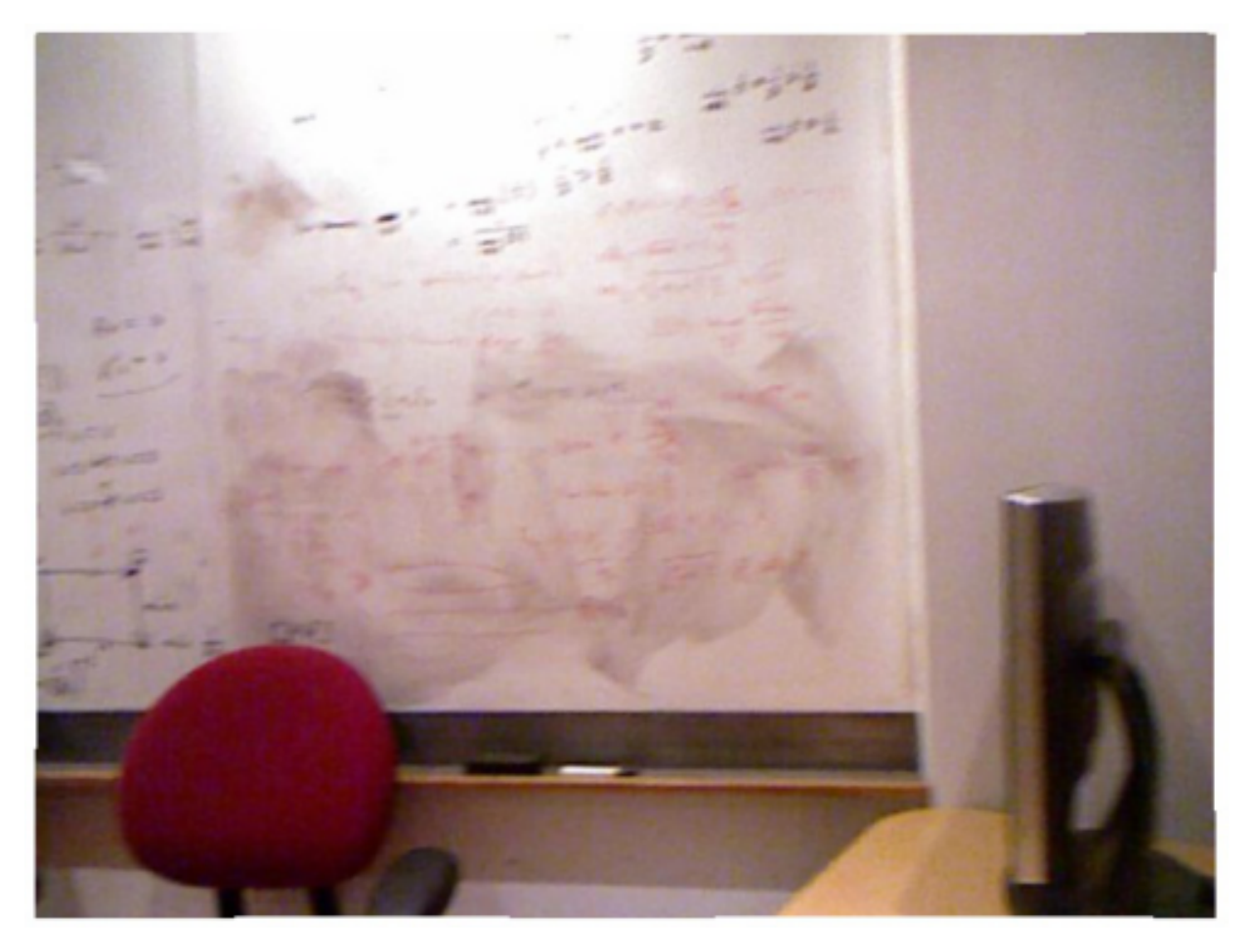} \\
\includegraphics[scale=0.2]{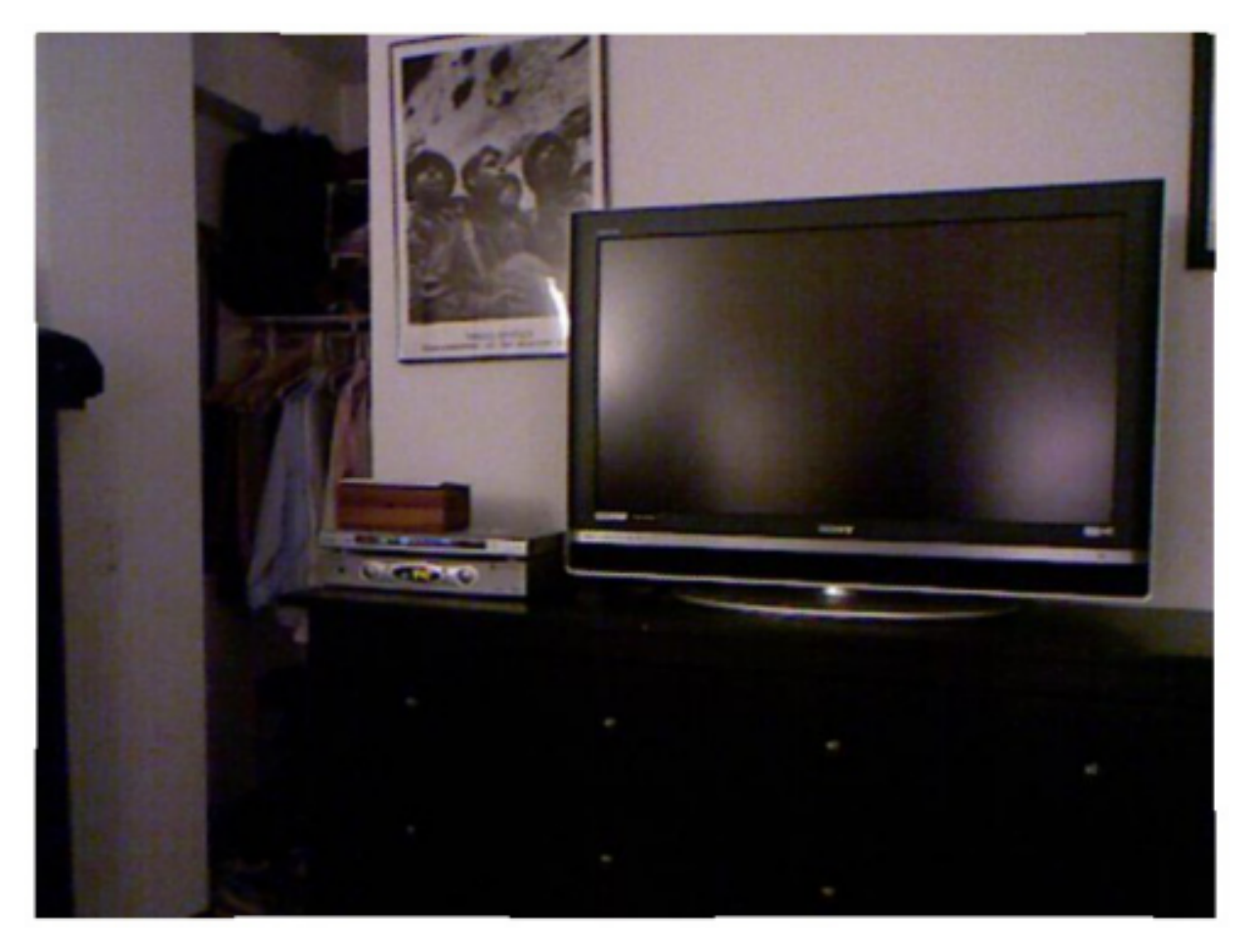} \\
\includegraphics[scale=0.2]{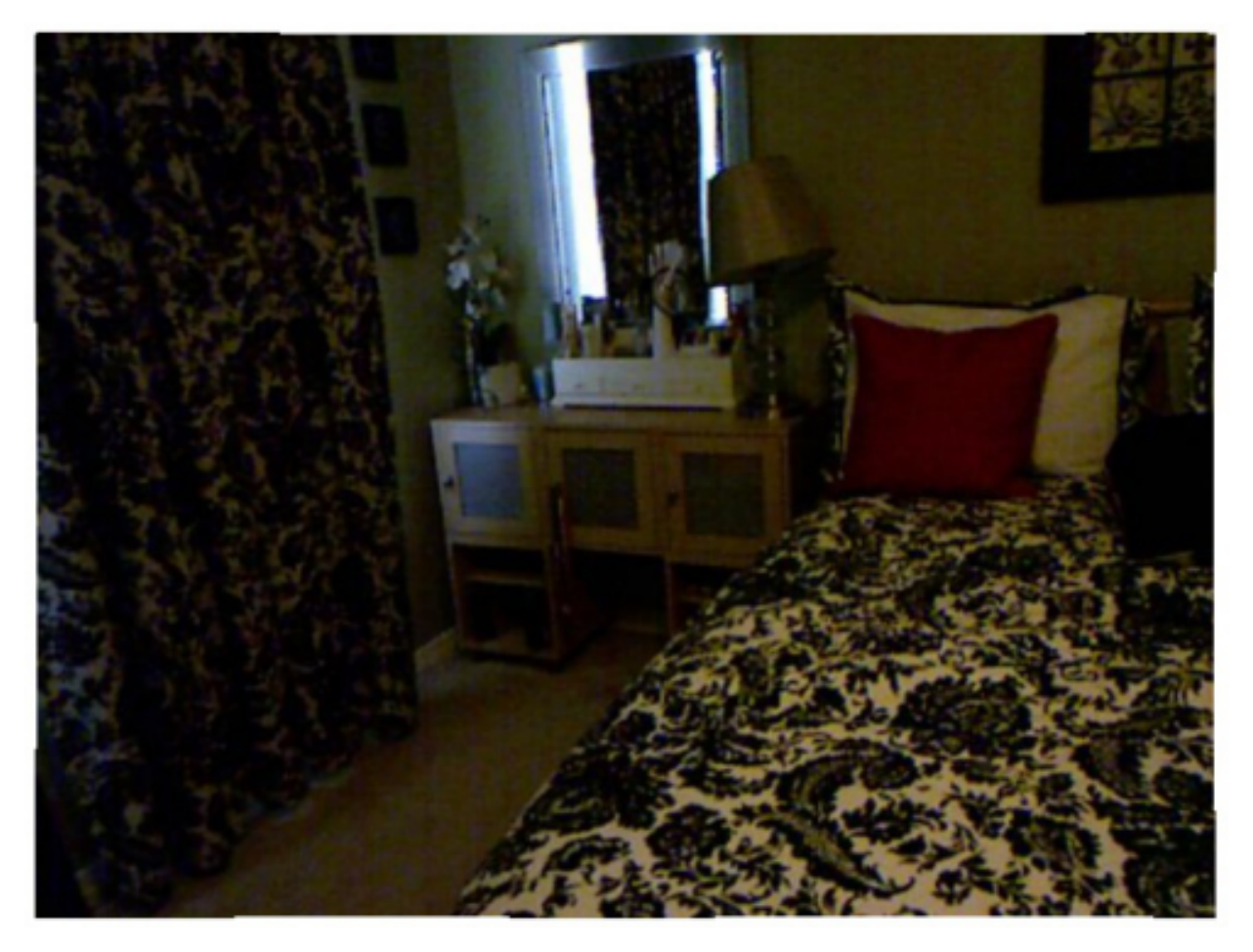} \\
\includegraphics[scale=0.2]{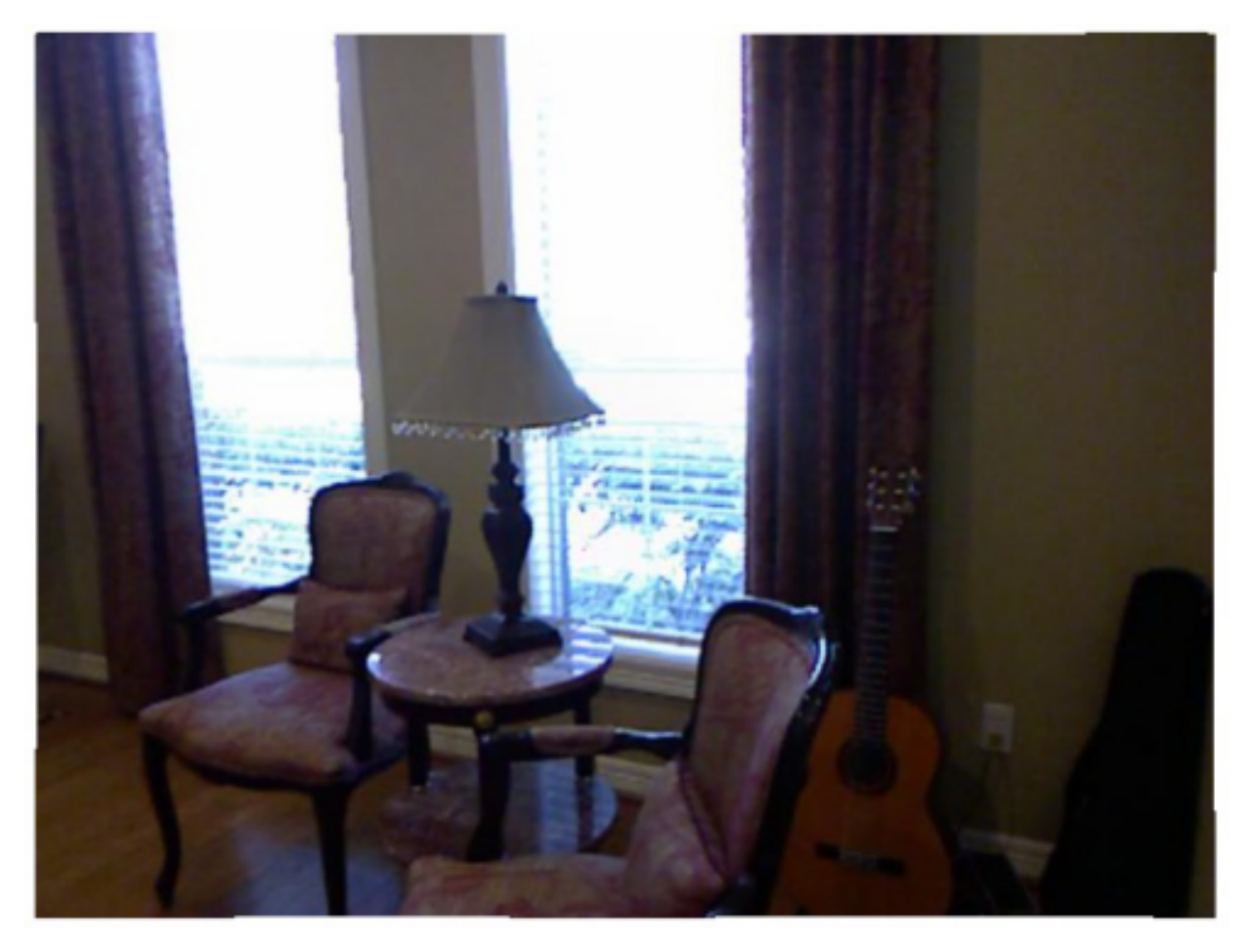}
\end{minipage}
}\hspace{-2.5mm}
\subfloat[depth images]{
\begin{minipage}[b]{0.15\textwidth}
\centering
\includegraphics[scale=0.2]{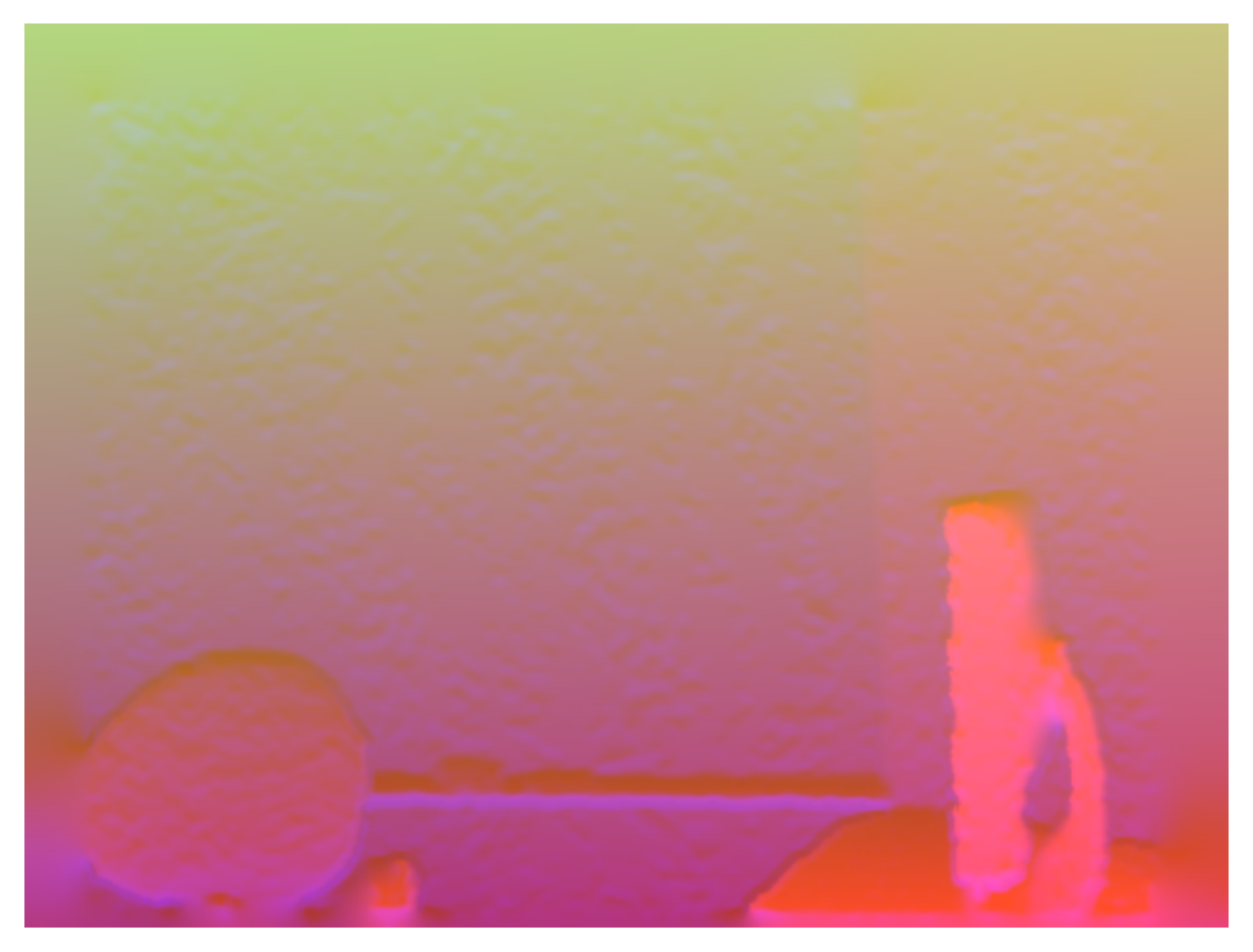} \\
\includegraphics[scale=0.2]{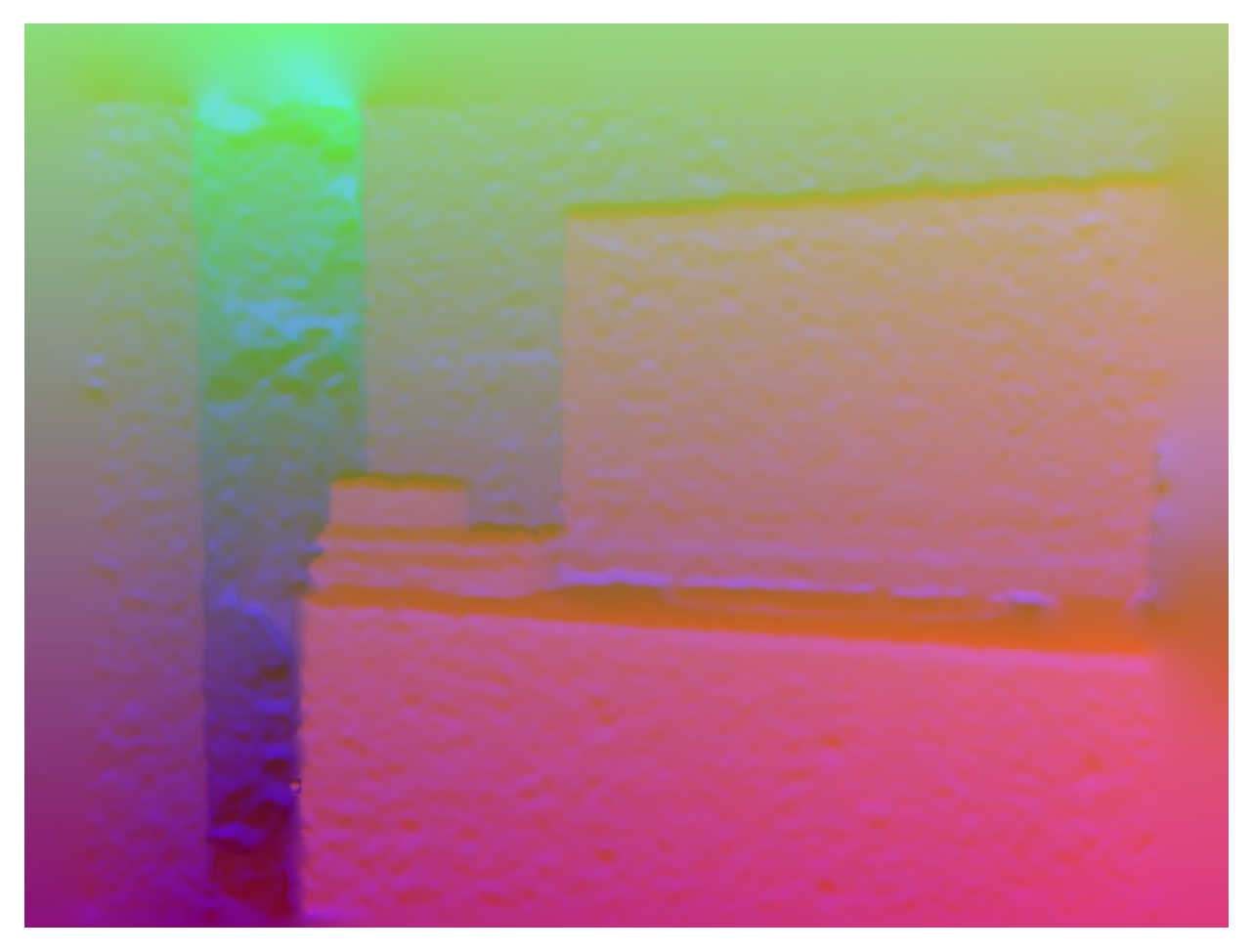} \\
\includegraphics[scale=0.2]{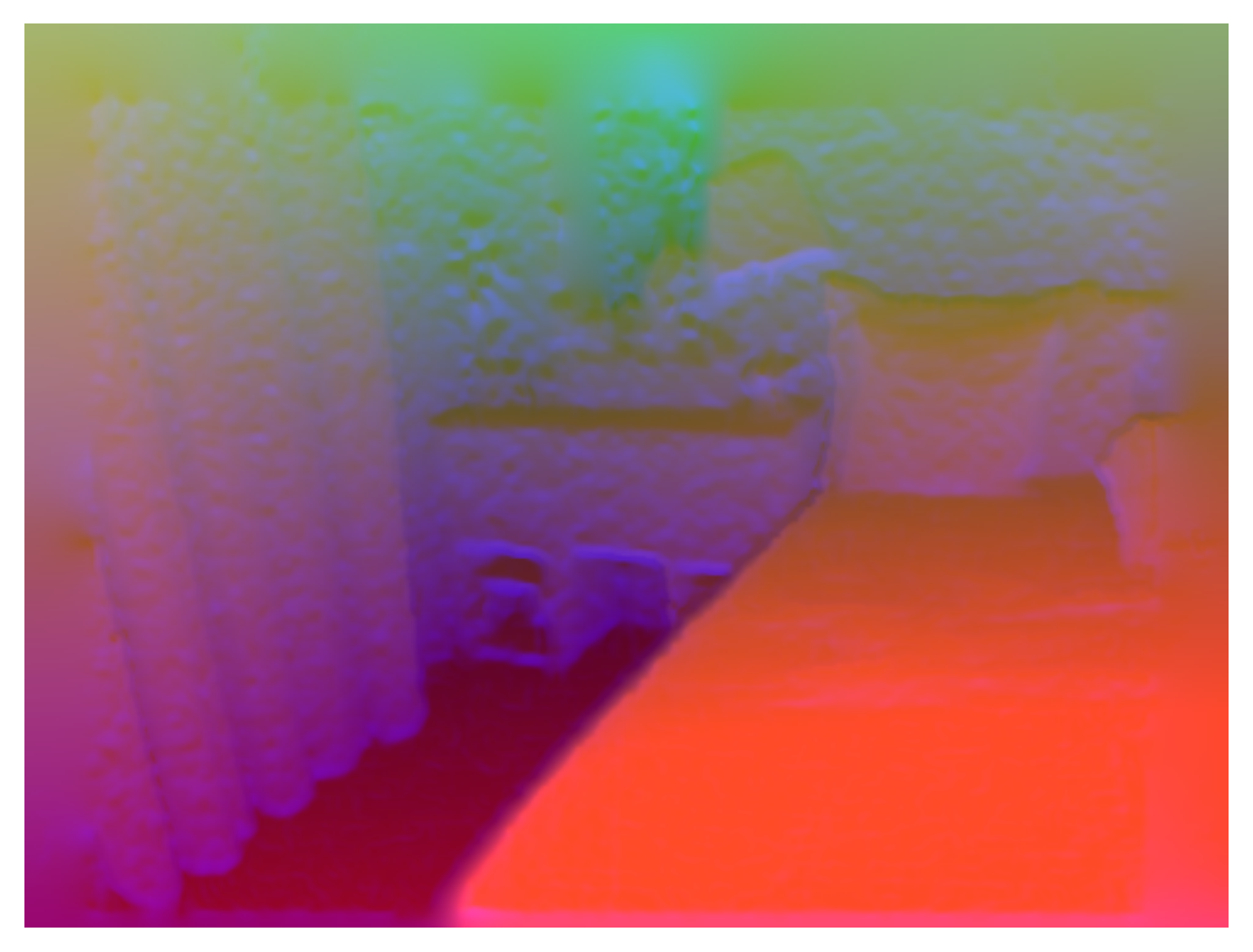} \\
\includegraphics[scale=0.2]{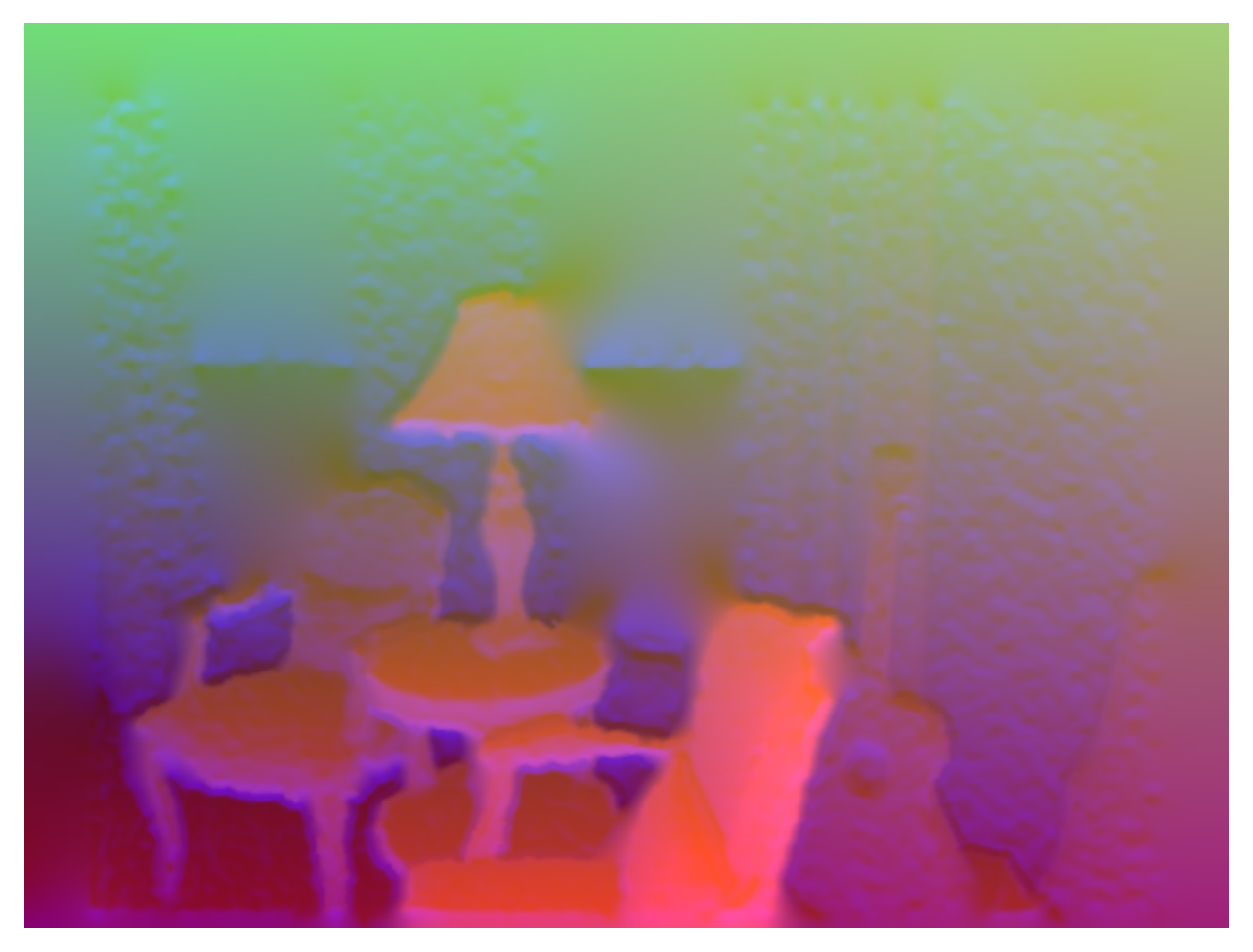}
\end{minipage}
}\hspace{-2.5mm}
\subfloat[ground truth]{
\begin{minipage}[b]{0.15\textwidth}
\centering
\includegraphics[scale=0.2]{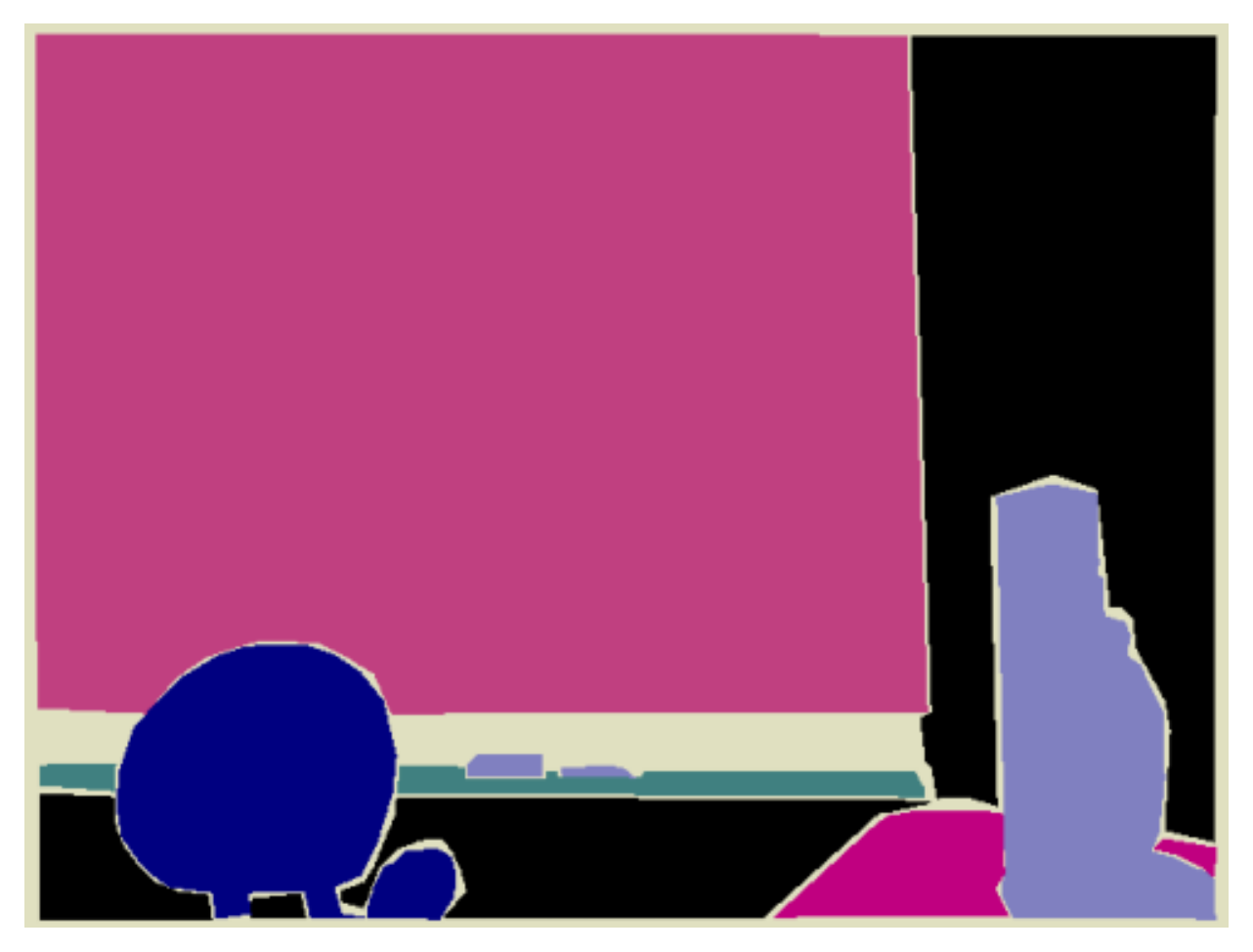} \\
\includegraphics[scale=0.2]{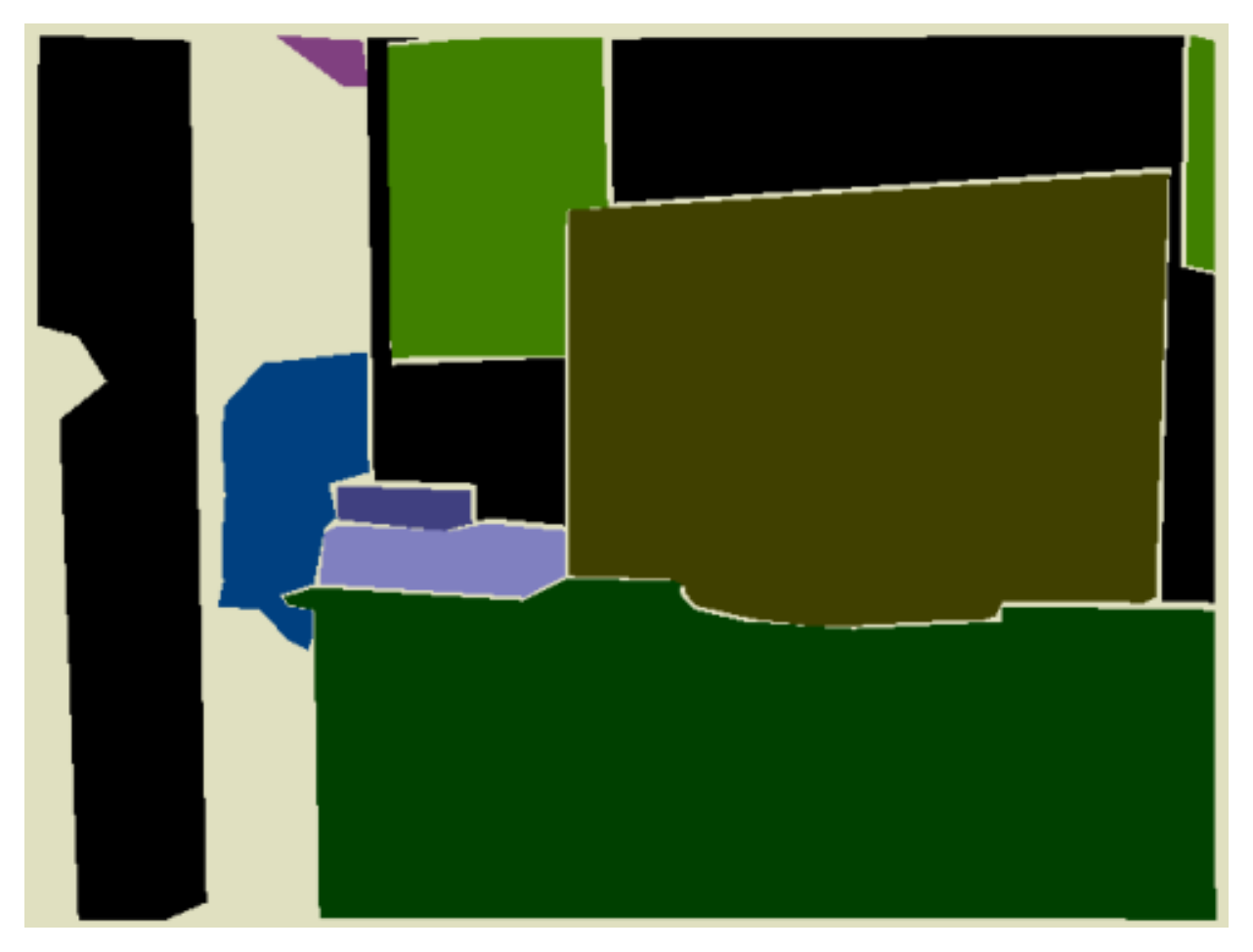} \\
\includegraphics[scale=0.2]{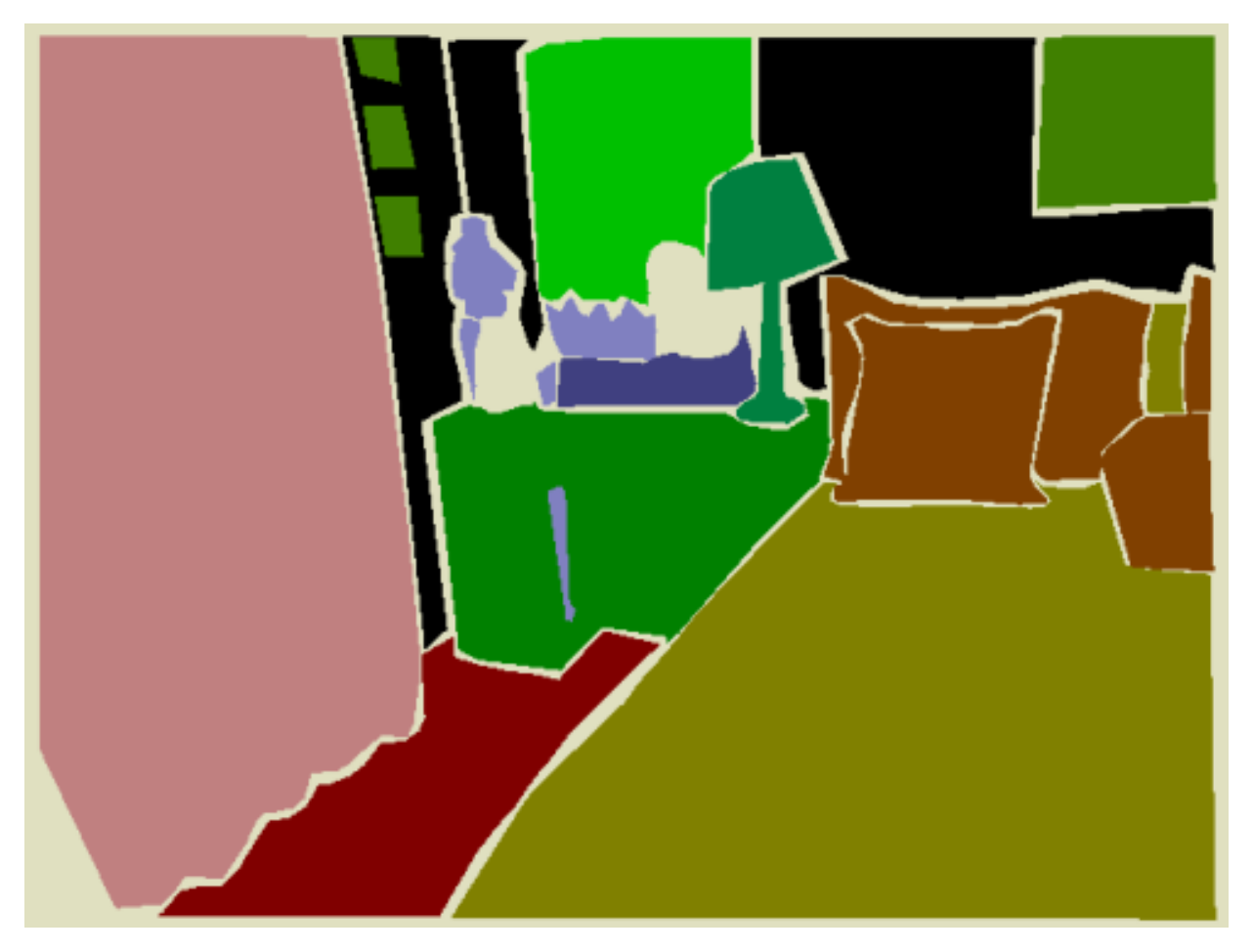} \\
\includegraphics[scale=0.2]{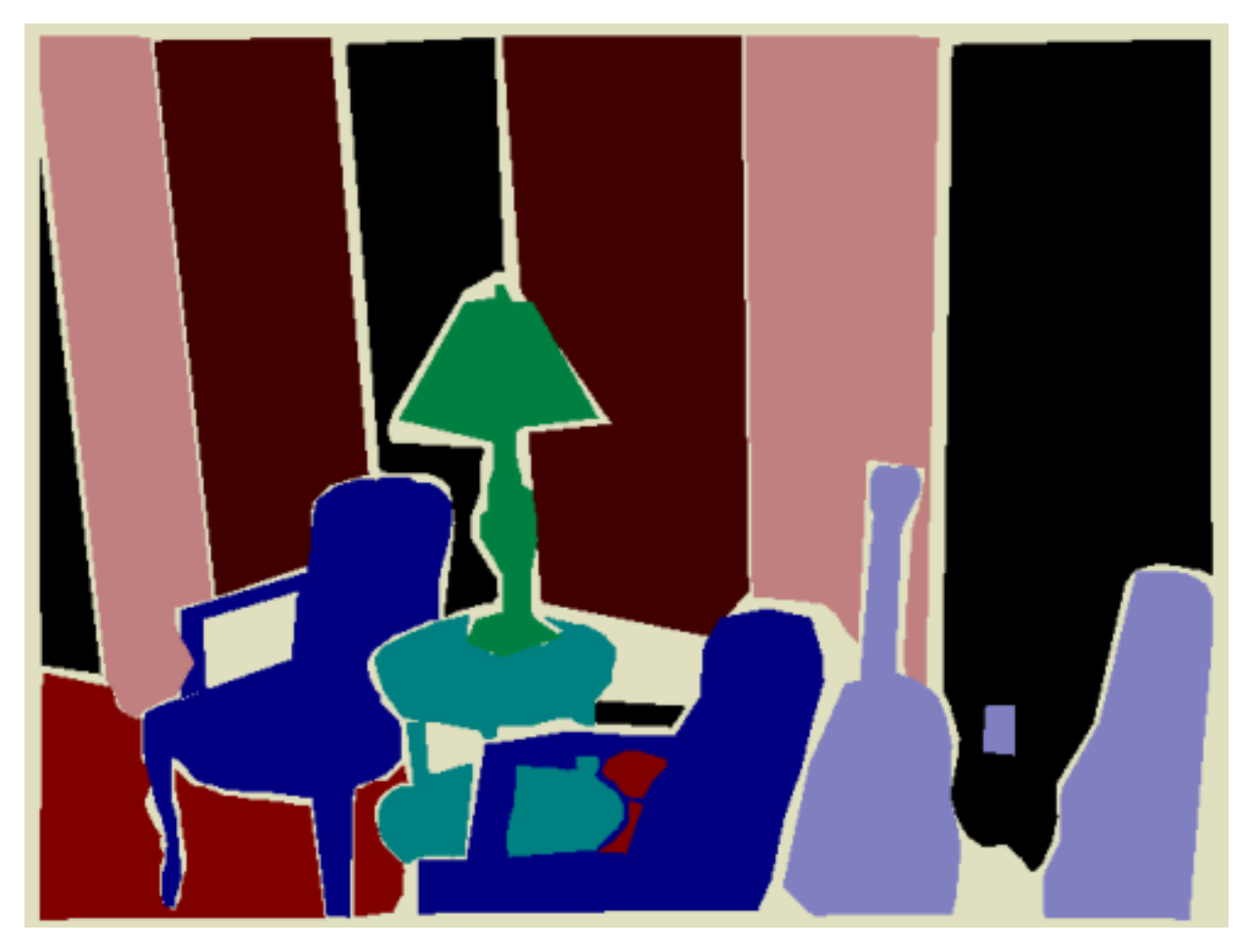}
\end{minipage}
}\hspace{-2.5mm}
\subfloat[MM student (ours)]{
\begin{minipage}[b]{0.15\textwidth}
\centering
\includegraphics[scale=0.2]{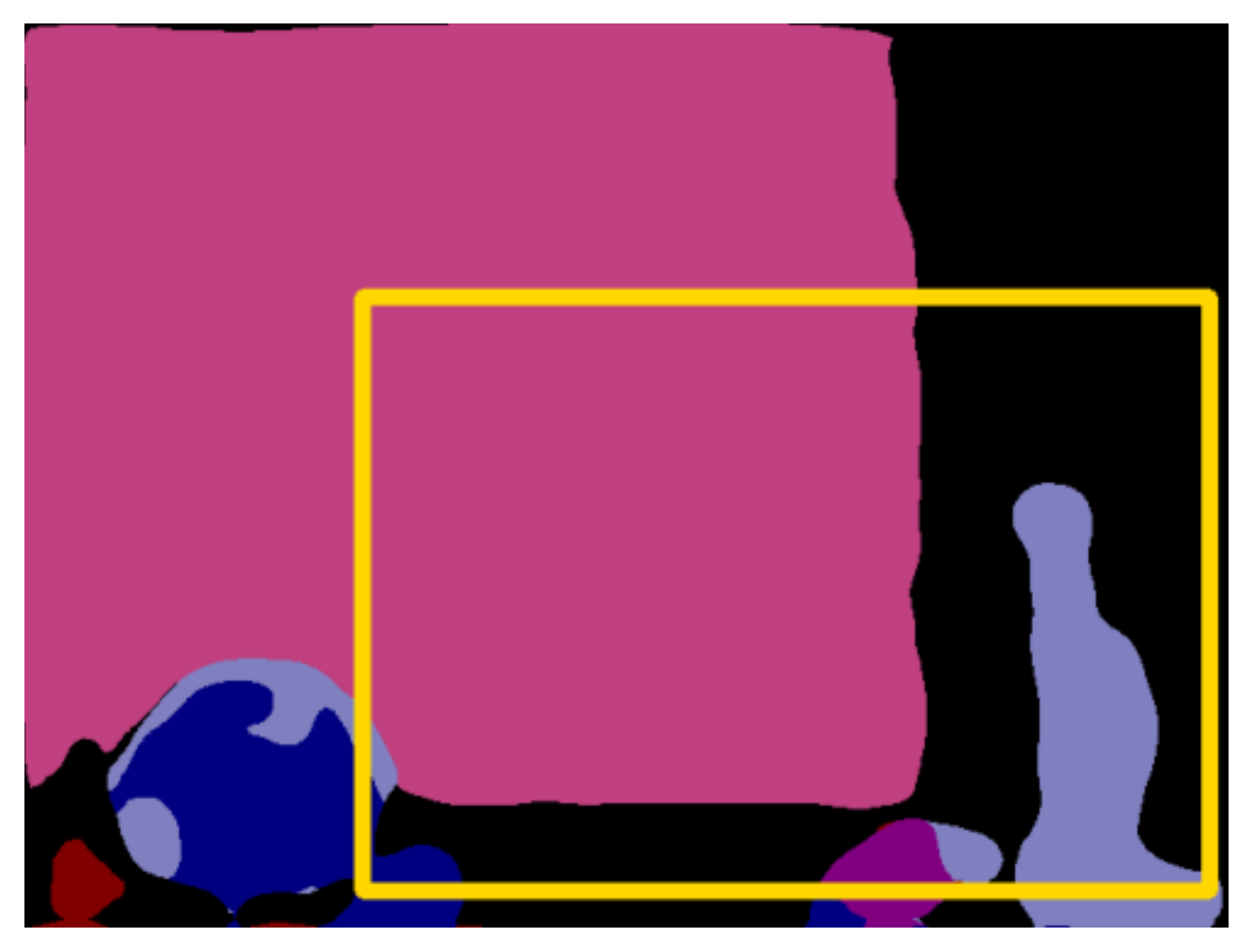} \\
\includegraphics[scale=0.2]{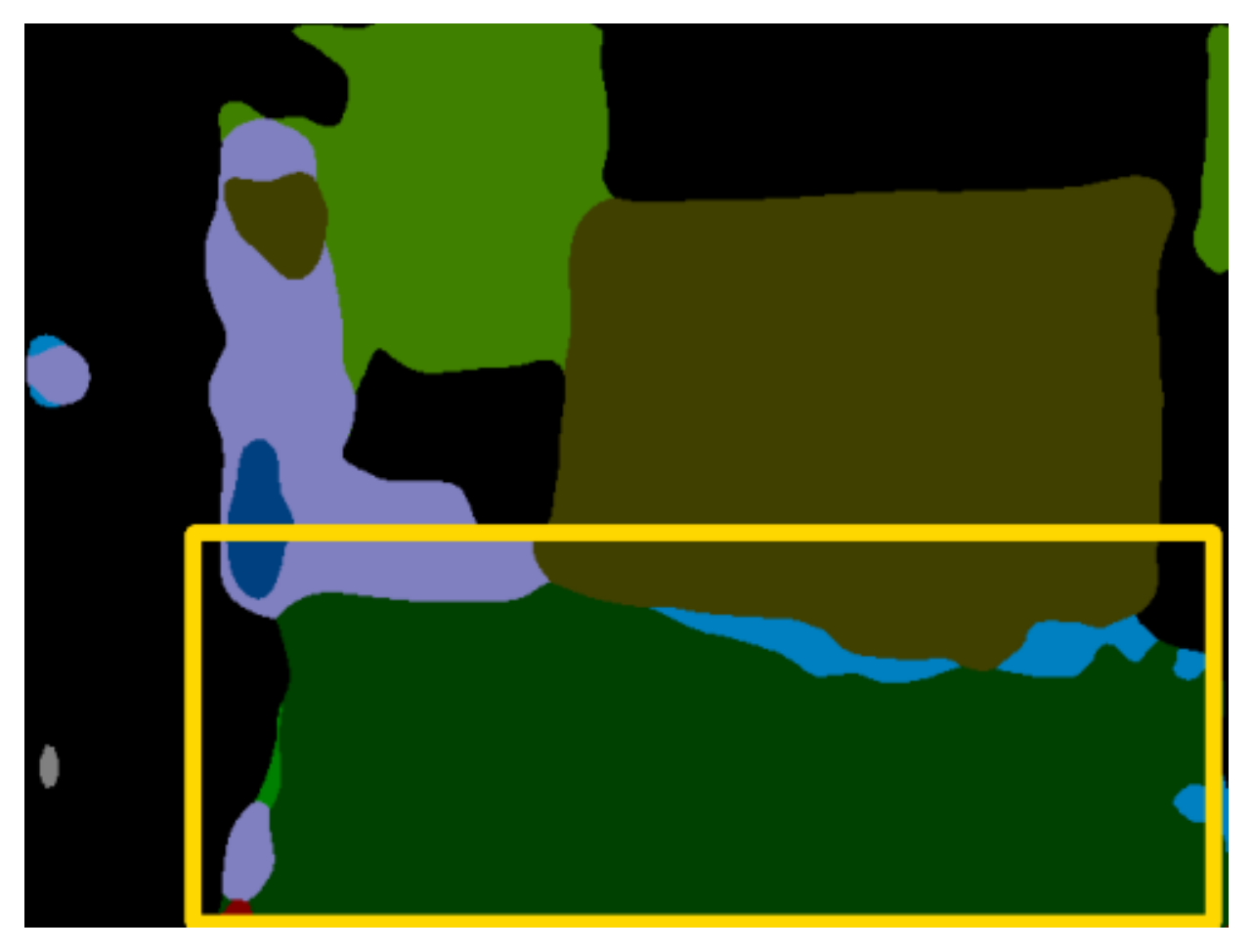} \\
\includegraphics[scale=0.2]{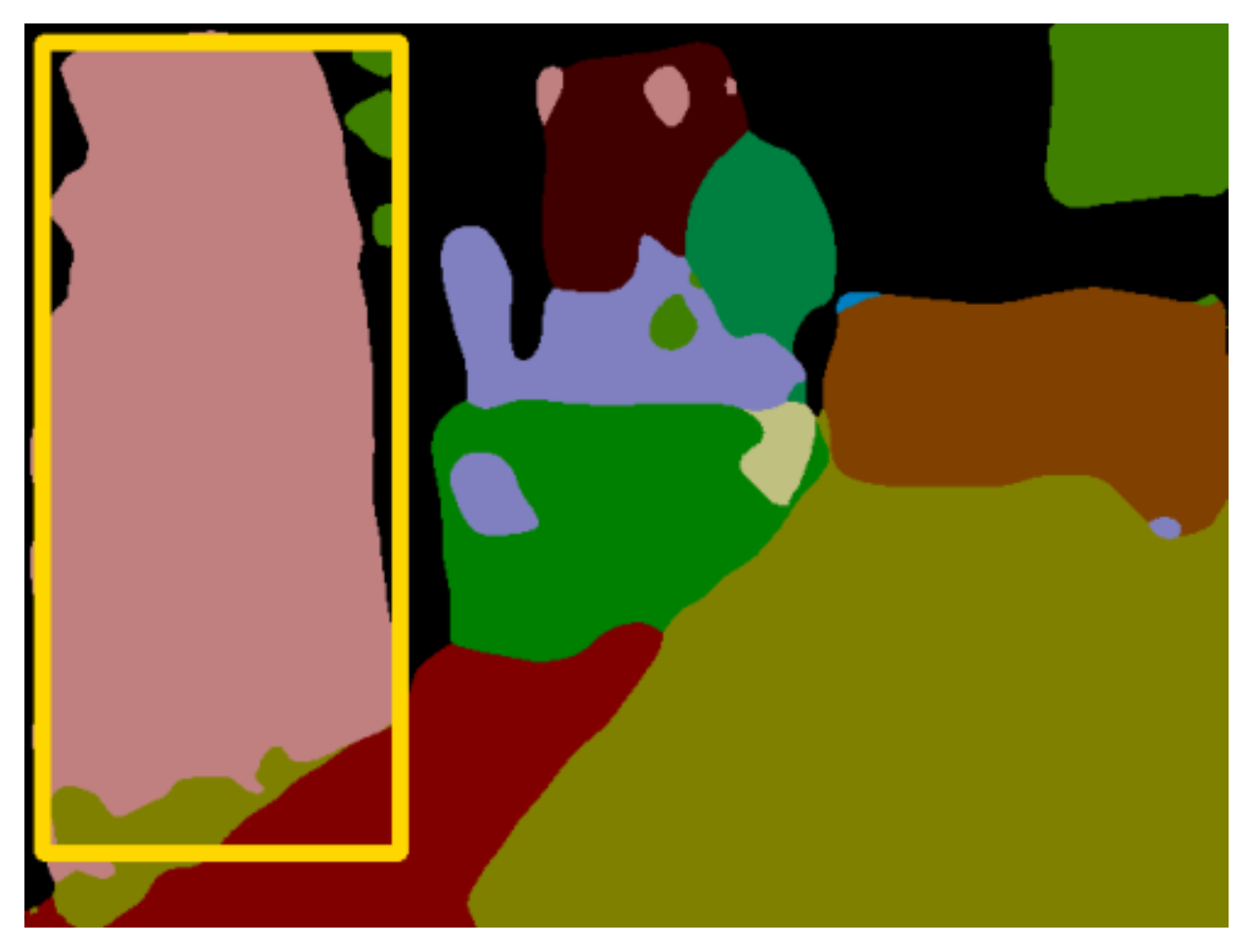} \\
\includegraphics[scale=0.2]{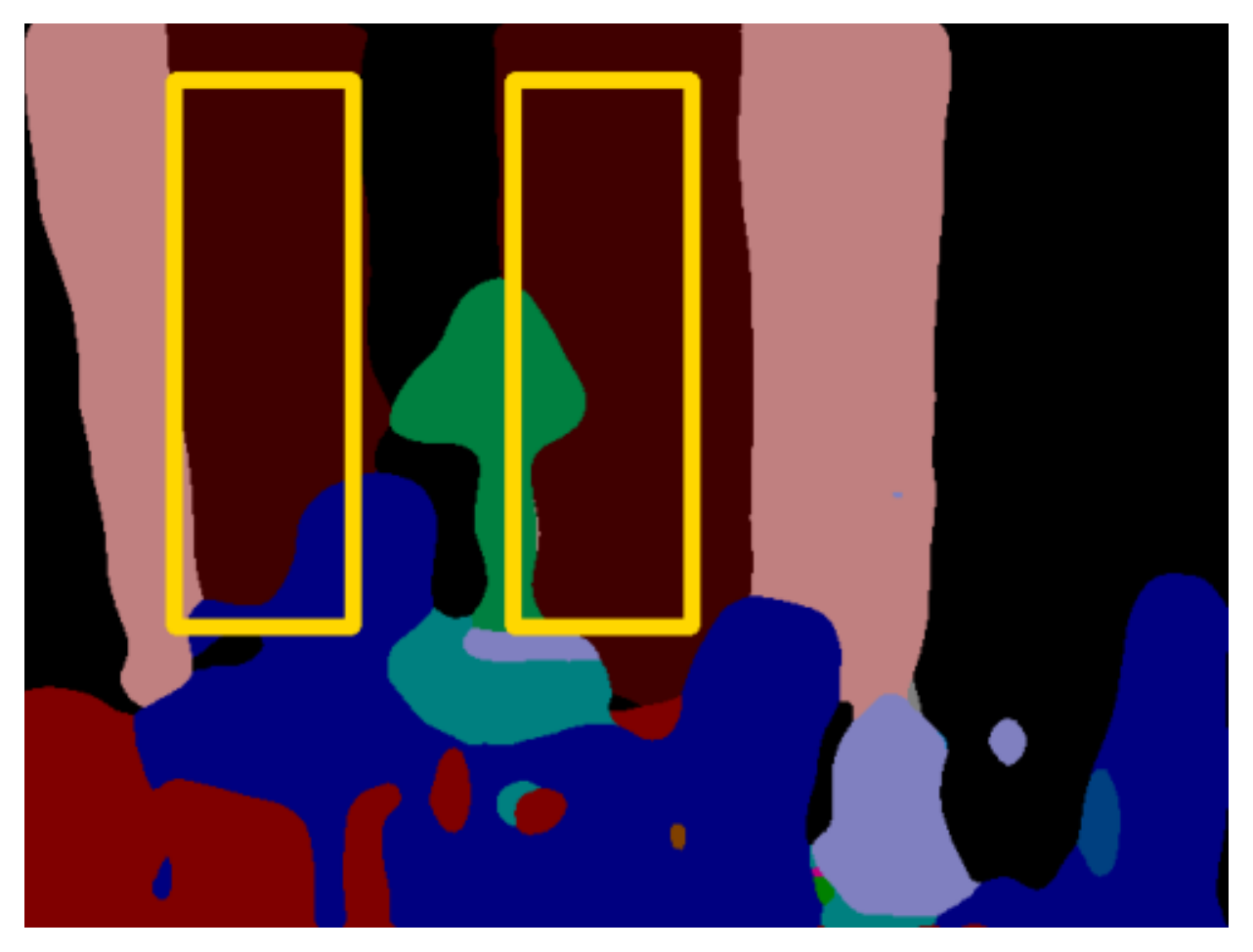}
\end{minipage}
}\hspace{-2.5mm}
\subfloat[UM teacher]{
\begin{minipage}[b]{0.15\textwidth}
\centering
\includegraphics[scale=0.2]{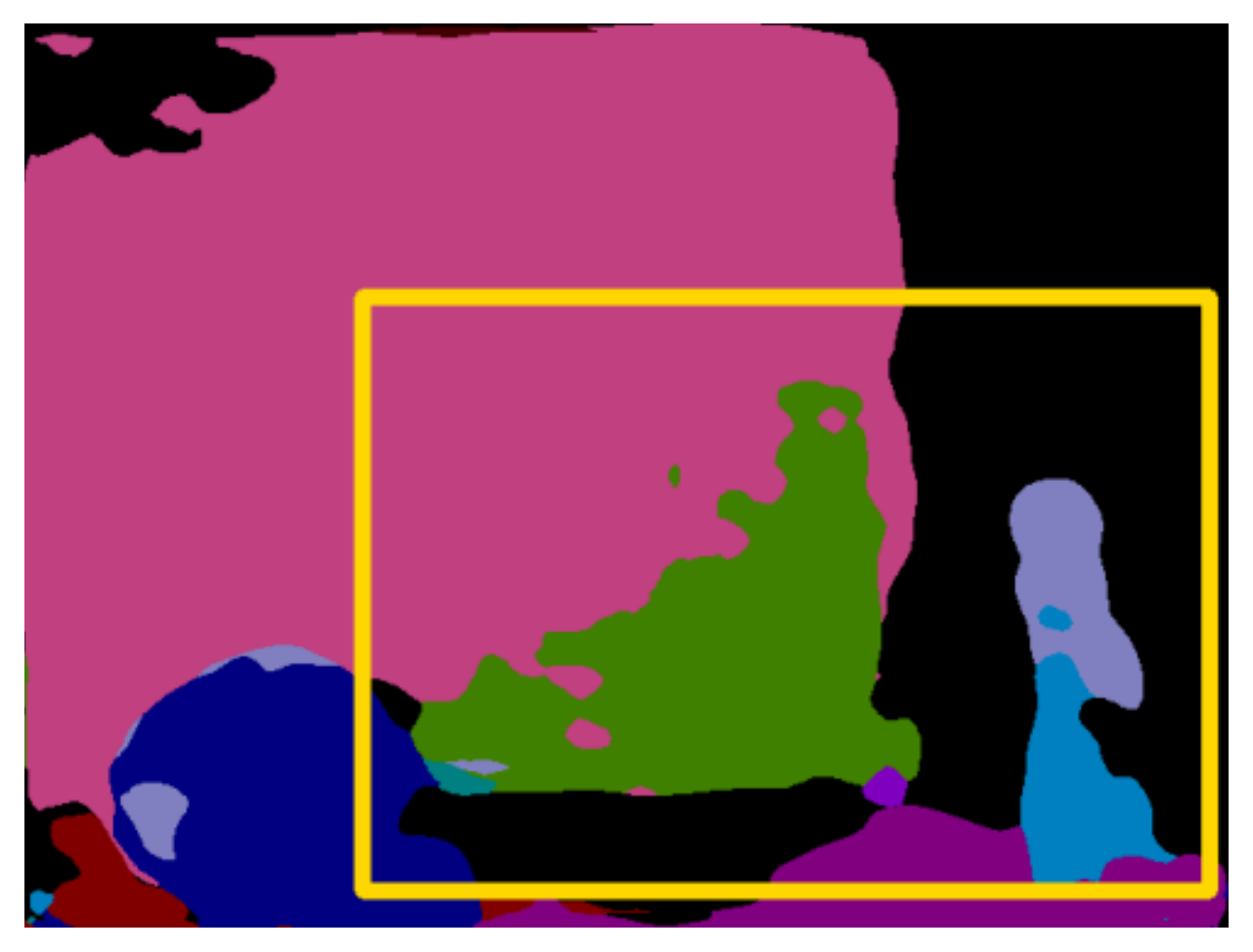} \\
\includegraphics[scale=0.2]{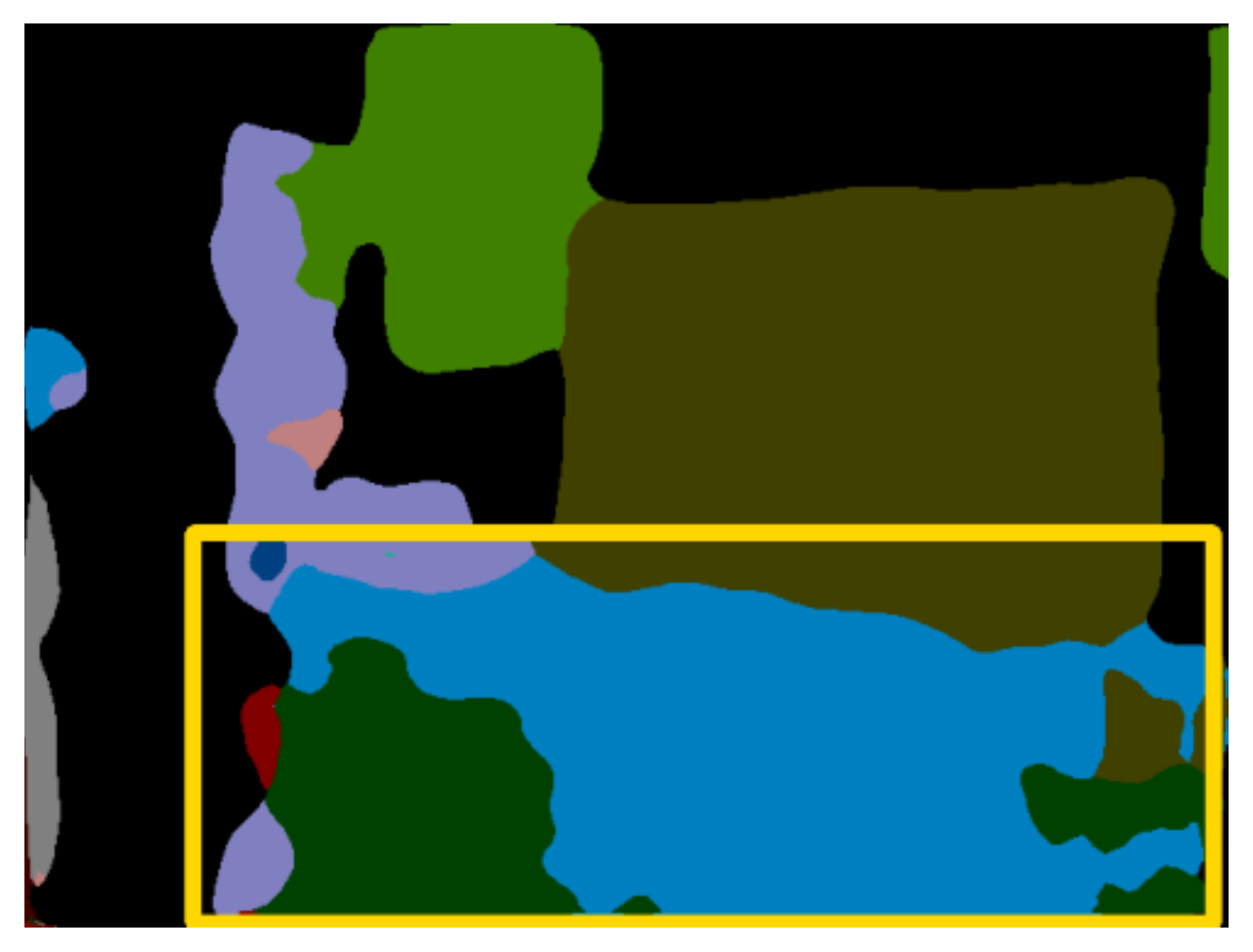} \\
\includegraphics[scale=0.2]{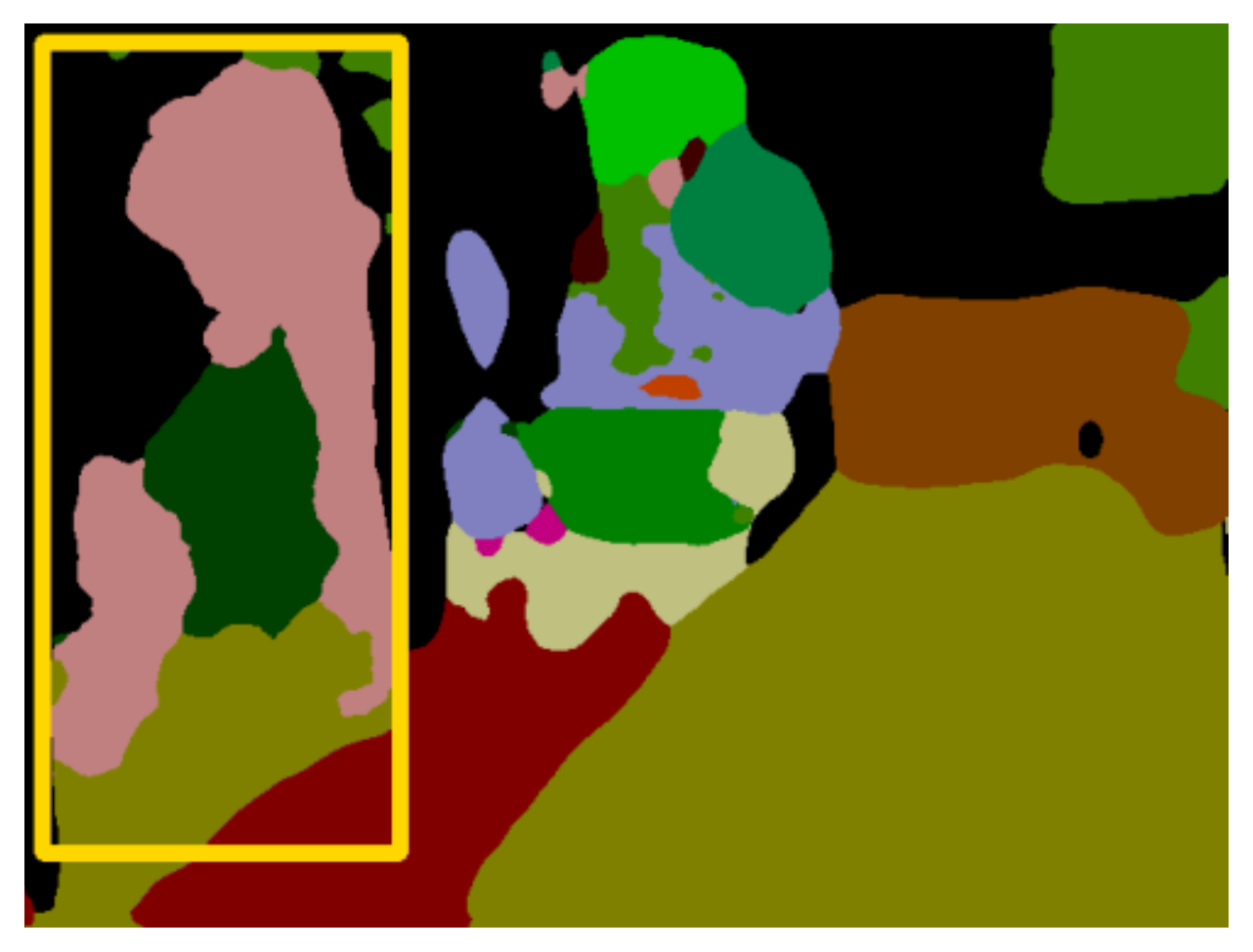} \\
\includegraphics[scale=0.2]{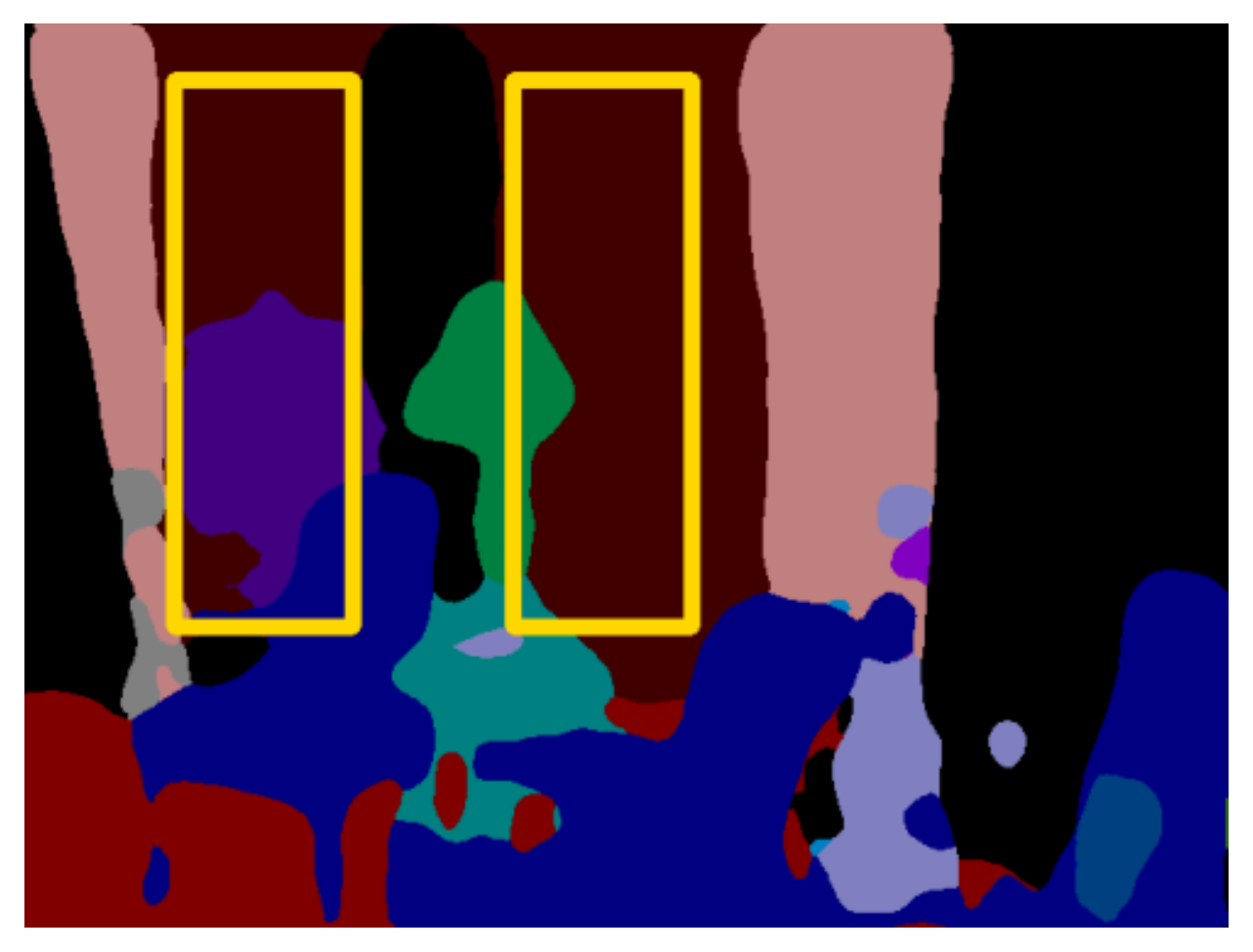}
\end{minipage}
}\hspace{-2.5mm}
\subfloat[NOISY student]{
\begin{minipage}[b]{0.15\textwidth}
\centering
\includegraphics[scale=0.2]{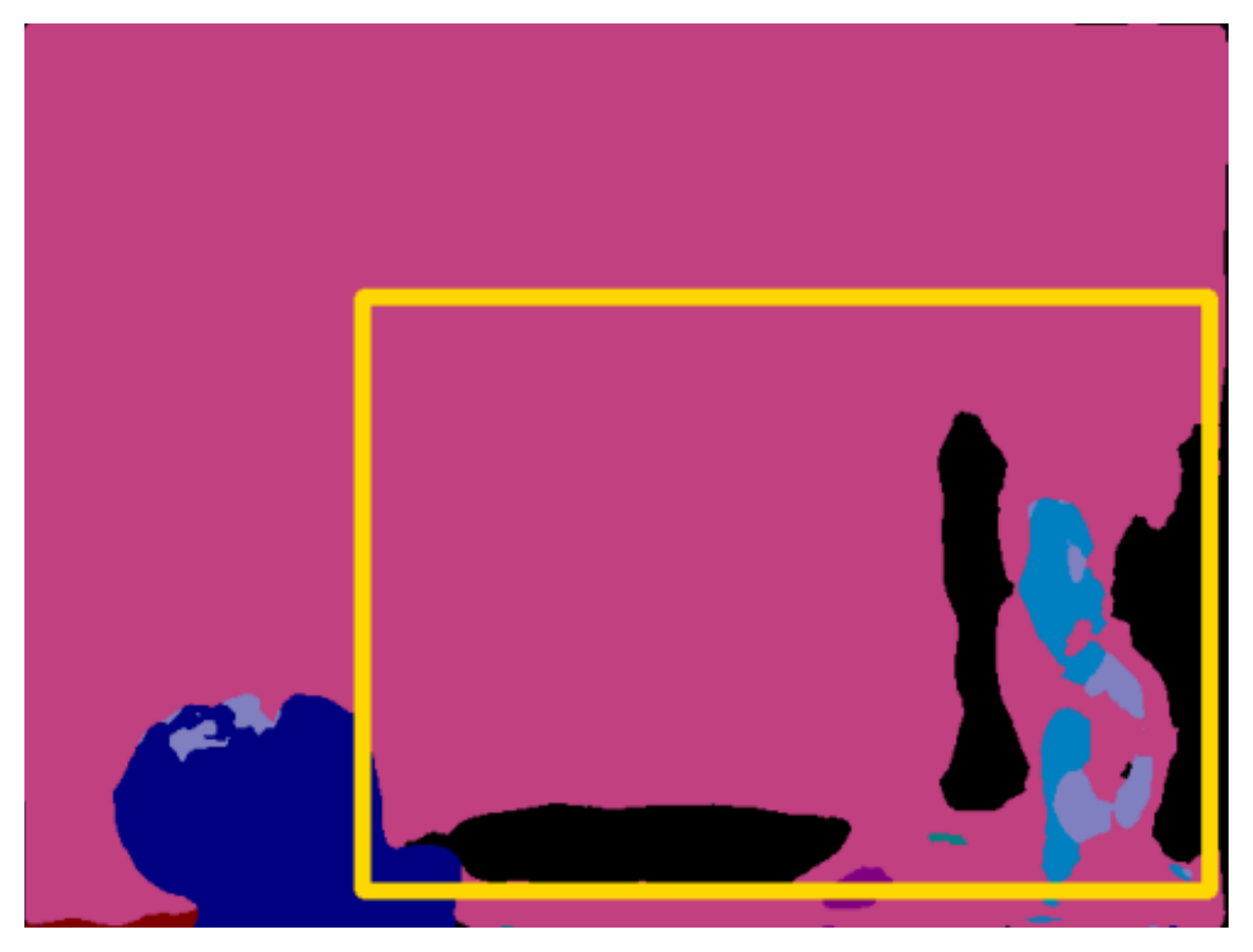} \\
\includegraphics[scale=0.2]{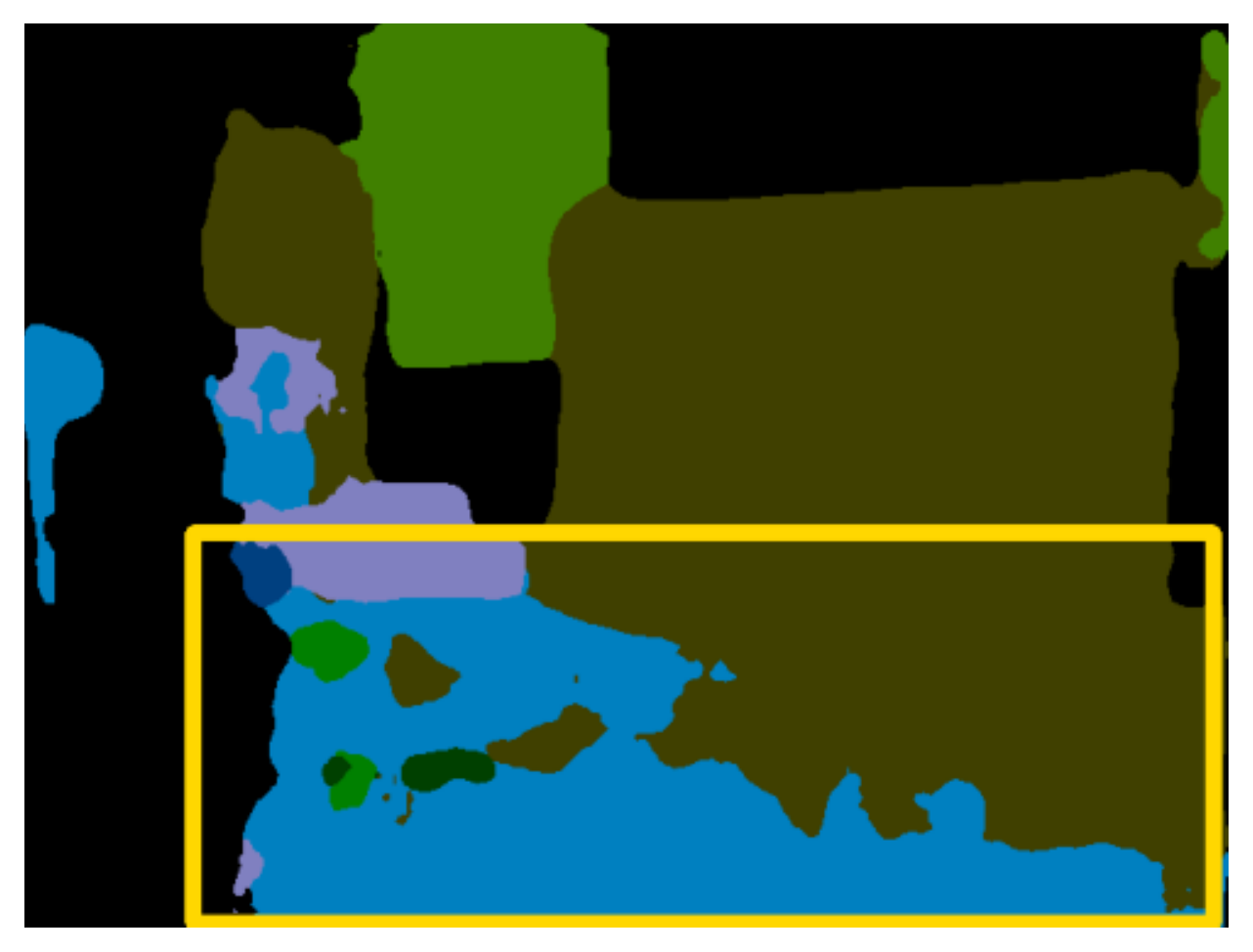} \\
\includegraphics[scale=0.2]{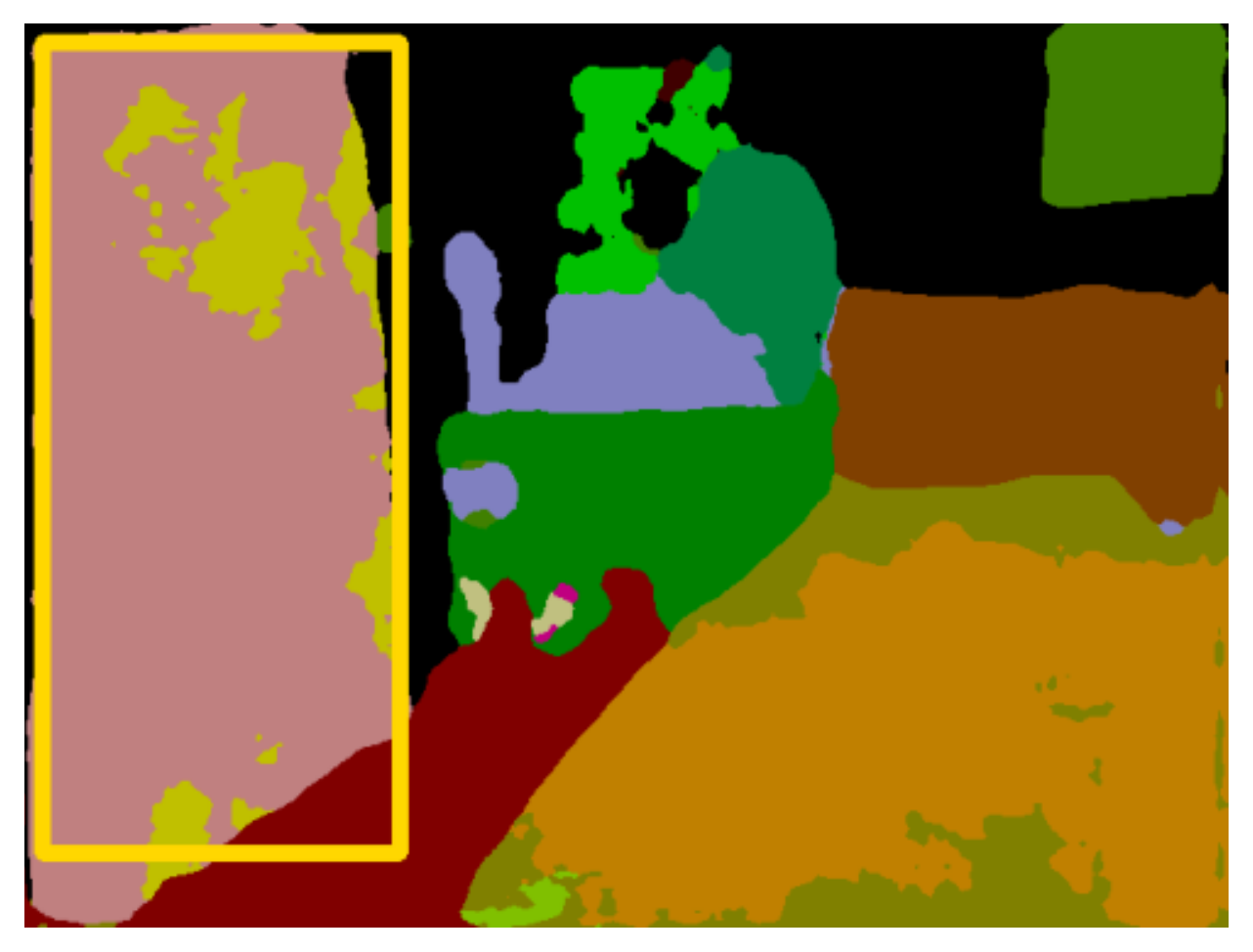} \\
\includegraphics[scale=0.2]{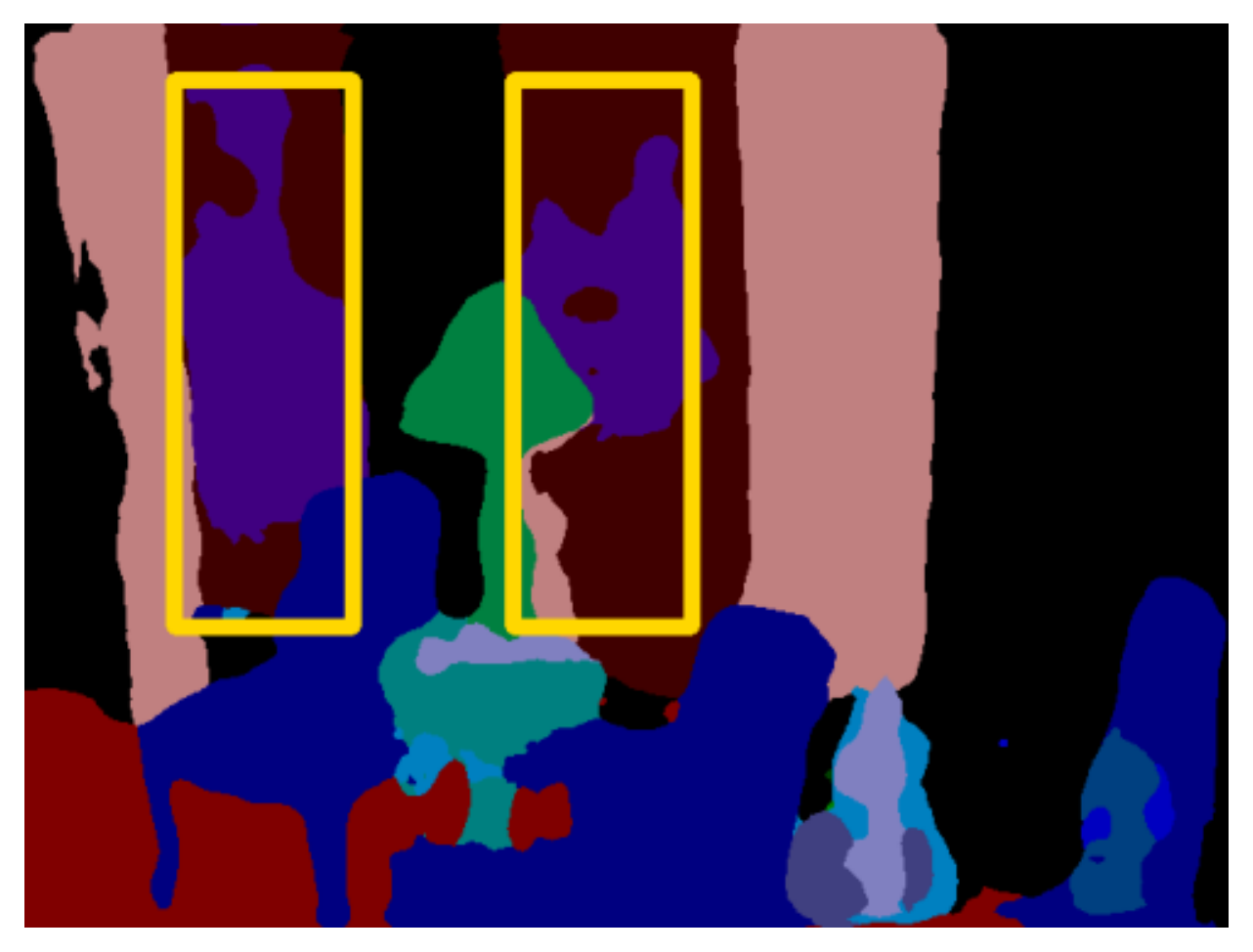}
\end{minipage}
}\hspace{-2.5mm}
\caption{Qualitative segmentation results on NYU Depth V2 test set.}
\label{fig:seg}
\end{figure*}
~\\
\noindent\textbf{Baselines \& Implementation.}
We compare \textit{MKE} against SSL methods \cite{naivestudent} \cite{noisystudent} and cross-modal distillation \cite{cmd} \cite{cmkd}. Due to different problem settings, we slightly modified cross-modal distillation methods to make them comparable. Since RGB-D images from $D_u$ are unannotated, we are unable to train a supervised version of the MM student (\ie, MM student (sup)) in this task.
We adopt ResNet-101 \cite{he2016deep} as backbone and DeepLab V3+ \cite{deeplabv3} as decoder for the UM teacher. In terms of training a MM student, depth images are first converted to HHA images and then passed to a fusion network architecture proposed in \cite{sagate} along with RGB images. We design the UM student architecture as the RGB branch of a MM student network. For the regularization term, we employ input augmentation for RGB images, \ie, random horizontal flipping and scaling with scales [0.5,1.75]. 
~\\

\noindent\textbf{Results.} Table \ref{tab:seg} reports mean Intersection-over-Union (mIoU) of each method. We observe that a MM student greatly improves over the UM teacher, \ie, achieves a mIoU of 48.88 \% while it is trained on pseudo labels of approximately 44.15\% mIoU. Furthermore, provided with no ground truth, our MM student outperforms all baselines with a considerable performance gain. This demonstrates the effectiveness of \textit{MKE}. We also arrive at the same conclusion that regularization helps improve the MM student since our MM student yields higher accuracy than a MM student (no reg). It indicates that \textit{MKE} and current SSL methods that focus on designing augmentations to emphasize consistency regularization can be combined together to boost performance.
~\\

\noindent Visualization results in Figure \ref{fig:seg} demonstrate that our MM student refines pseudo labels and achieves knowledge expansion. Although it receives noisy predictions given by the UM teacher, our MM student does a good job in handling details and maintaining intra-class consistency. As shown in the third and fourth row, the MM student is robust to illumination changes while the UM teacher and NOISY student easily get confused. Depth modality helps our MM student better distinguish objects and correct wrong predictions it receives. More qualitative examples are shown in the supplementary material.

\subsection{Event Classification}

% \begin{table}[h]\scriptsize\vspace{-3mm}
% 	\renewcommand\tabcolsep{1.5pt}
	
% 	\centering
% 	%	\setlength{\tabcolsep}{3mm}{
	
% 	\resizebox{0.48\textwidth}{!}{
% 	\begin{tabular}{c ||c|c|c}
% 		\toprule		
		
% 		Method&'basketball bounce’&'dog growling‘&'people belly laughing'\\
% 		\midrule
% 		UM teacher&0.178&0.069&0.334\\
% 	    MM student (ours)&0.542&0.516&0.8\\
		
% 		\bottomrule
% 	\end{tabular}}
% 	\label{tab:misrec}
% 	\caption{The top x events with average precision improvements in our method}
% \end{table}

We present experimental results on a real-world application, event classification. 3.7K audios from AudioSet~\cite{audioset} and 3.7K audio-video pairs from VGGSound~\cite{vggsound} are taken as the labeled unimodal dataset $D_l$ and unlabeled multimodal dataset $D_u$, respectively. In this task, modality $\alpha$ and $\beta$ correspond to audios and videos.
~\\

% In this experiment, audio plays the main role and video introduces additional information. We consider audio as a 'strong' modality $\alpha$ and video as a 'weak' modality $\beta$.

% \textbf{Dataset.} The AudioSet~\cite{} and VGGSound~\cite{} are both audio-visual datasets for event classification. We take a mini common set of them including 3710 data in AudioSet and 3748 data for training and 1937 data for testing in VGGSound with 46 event categories. VGGSound guarantees the audio-video correspondence that the sound source is visually evident within the video, while AudioSet does not. Therefore, we consider AudioSet as a unimodal dataset and VGG Sound as multimodal. Audios from AudioSet and audio-video pairs from VGGSound are taken as the labeled unimodal dataset $D_l$ and unlabeled multimodal data $D_u$ respectively.

%To train our UM teacher, and the pretrained UM teacher generate pseudo labels for the audio-video pairs from VGGSound dataset. Due to the paired audio and video in VGGSound, our MM student is trained on the audios, videos from VGGSound and the pseudo labels. For comparison, we also train a UM student on the audio from VGGSound and the pseudo label. The noisy student indicate train on audio from AudioSet with GT and audio from VGGSound with pseudo label

\noindent\textbf{Baslines \& Implementation.}
For the UM teacher, we take ResNet-18 as the backbone and a linear layer as classification layer. For the MM student, the audio backbone is identical to that of the UM teacher, and the video backbone is a ResNet-18 with 3D convolution layers. Features from the audio and video backbone are concatenated together before feeding into one classification layer. Following the same regularization term of \cite{vggsound}, we randomly sample audio clips of 5 seconds and apply short-time Fourier Transformation for 257 $\times$ 500 spectrograms during training. 
~\\

\noindent\textbf{Results.} Table \ref{Tab:AV} reports mean Average Precision (mAP) of each method. The baseline model is the UM teacher trained on $D_l$, which achieves a 0.345 mAP.  Benefiting from the video modality, our MM student achieves best performance with a mAP of 0.427, outperforming NOSIY student \cite{noisystudent} and cross-modal distillation methods \cite{owens2016ambient} \cite{cmkd}. Notably, the difference between our MM student and its upper bound (\ie, MM student (sup)) is small, showing great potentials of \textit{MKE} in correcting pseudo labels.

\renewcommand\arraystretch{1.0}
\begin{table}[!htbp]
\centering
\label{tab:avset}
\begin{tabular}{ccccc}

\toprule
\multirow{2}*{Method} & \multicolumn{3}{c}{Train data} & \multirow{2}*{\makecell[c]{Test mAP}} \\
~ & \textit{mod} & $D_l$ & $\tilde{D}_u$ &  ~ \\
\midrule
UM teacher & $a$ & $\checkmark$ & &  0.345 \\
UM student & $a$ &  &$\checkmark$  &  0.406 \\
NOISY student \cite{noisystudent}& $a$ & $\checkmark$ & $\checkmark$ & 0.411 \\
Owens \etal~\cite{owens2016ambient} & $a, v$ & & $\checkmark$ & 0.371\\
CMKD~\cite{cmkd} & $a, v$ & & $\checkmark$ & 0.372\\
MM student (no reg) & $a, v$ & & $\checkmark$ & 0.421\\
MM student (ours)& $a, v$ & & $\checkmark$ & \textbf{0.427}\\
\hdashline
%\midrule
\makecell[c]{MM student (sup)} & $a, v$ &  & $\star$ & 0.434\\
\bottomrule

\end{tabular}
\caption{Results of event classification on AudioSet and VGGSound. $a$ and $v$ indicate audios and videos.}
\label{Tab:AV}
\end{table}

% Due to space limit, we present a comprehensive study of various components of \textit{MKE}, \ie, unlabeled data size, teacher model, hard / soft pseudo labels in the supplementary material. In addition, dataset and implementation details, and more qualitative examples on emotion recognition, semantic segmentation and event classification are provided.

% \textbf{Results} As illustrated in Table \ref{Tab:AV}, the baseline model is the UM teacher trained on AudioSet data $D_l$, which achieves a 0.345 mAP. With pseudo labels on VGGSound ($D_u$) generated from UM teacher, we train the lower bound model, UM student on the audio from VGGSound and the pseudo label from UM teacher. Recent method, Noisy student~\cite{}, aims at improving the performance with iterative training student and teacher, therefore noisy student outperform the lower bound method (UM student) and achieve 0.005 mAP improvement. In contrast, our method introduce extra modality (video modality) as counterpart information and achieves 0.425 mAP (state-of-the-art performance) . For comparison, we train a supervised MM student (MM student (sup)) as the upper bound of our methods with audio, video and label of VGGSound. Our MM student trained on pseudo labels achieves comparable performance with the MM student (sup) trained on true labeles of $D_u$, showing great potentials of \textit{MKE} in correcting pseudo labels.

\section{Conclusion}

Motivated by recent progress on multimodal data collection, we propose a multimodal knowledge expansion framework to effectively utilize abundant unlabeled multimodal data. We provide theoretical analysis and conduct extensive experiments, demonstrating that a multimodal student corrects inaccurate predictions and achieves knowledge expansion from the unimodal teacher. In addition, compared with current semi-supervised learning methods, \textit{MKE} offers a novel angle in addressing confirmation bias.

% to address confirmation bias and obtain a student model with good representation power, we can resort to abundant unlabeled multimodal data in the input level
% \section{Problem}
% add: labeled data and unlabeled data do not overlap; orthogonal to noisy student (regularization)
% teacher model is not accessible.
% MM student is a good starting point.
% title

{\small
\bibliographystyle{ieee_fullname}
\bibliography{egbib}
}

% %%%%%%%%% SUPPLEMENTARY
% \newpage
% ~
% \newpage

% \twocolumn[
% \begin{@twocolumnfalse}
% \section*{\centering{\Large{Supplementary Material\\[12pt]}}}
% \end{@twocolumnfalse}
% ]
% % \begin{center}
% %     \Large {Supplementary Materials}
% % \end{center}

% This supplementary material presents: (1) dataset and implementation details; (2) more qualitative experimental results; (3) ablation studies; (4) proofs in Section 3.\\
% ~

% \input{supp1-data}
% \input{supp2-exp}
% \input{supp3-ablation}
% \input{supp4-proof}

\end{document}